\documentclass[lettersize,journal]{IEEEtran}
\usepackage{amsmath,amsfonts}
\usepackage{algorithmic}
\usepackage{algorithm}
\usepackage{array}
\usepackage[caption=false,font=normalsize,labelfont=sf,textfont=sf]{subfig}
\usepackage{textcomp}
\usepackage{stfloats}
\usepackage{url}
\usepackage{verbatim}
\usepackage{graphicx}
\usepackage{cite}
\usepackage{multirow}
\usepackage{booktabs}
\usepackage[pagebackref=false,breaklinks=true,colorlinks,bookmarks=false]{hyperref}

\usepackage{url}
\usepackage[dvipsnames]{xcolor}
\hypersetup{
	final,
	breaklinks,
	colorlinks,
	linkcolor={Mahogany},
	citecolor={YellowGreen},
	urlcolor={CadetBlue}
}

\usepackage[dvipsnames]{xcolor}

\begin{document}

\title{Surface-SOS: Self-Supervised Object Segmentation via Neural Surface Representation}
\author{Xiaoyun Zheng, Liwei Liao, Jianbo Jiao,~\IEEEmembership{Member, IEEE}, Feng Gao, Ronggang Wang,~\IEEEmembership{Member, IEEE}
\thanks{Xiaoyun Zheng, Liwei Liao, Ronggang Wang are with School of Electronic and Computer Engineering, Peking University Shenzhen Graduate School, Shenzhen 518055, China, and also with Peng Cheng Laboratory, Shenzhen 518000, China (Email: xyun\_z@stu.pku.edu.cn; levio@pku.edu.cn; rgwang@pkusz.edu.cn).}
\thanks{Jianbo Jiao is with School of Computer Science, University of Birmingham, Edgbaston, Birmingham, B15 2TT, United Kingdom (Email: j.jiao@bham.ac.uk).}
\thanks{Feng Gao is with School of Arts, Peking University, Beijing 100871, China (Email: gaof@pku.edu.cn).}
\thanks{Corresponding author: Ronggang Wang.}
\thanks{Manuscript received July 14, 2023; revised January 9, 2024; accepted February 21, 2024.}
}

\markboth{Journal of \LaTeX\ Class Files,~Vol.~14, No.~8, August~2021}%
{Shell \MakeLowercase{\textit{et al.}}: A Sample Article Using IEEEtran.cls for IEEE Journals}


 \maketitle

\begin{abstract}
Self-supervised Object Segmentation (SOS) aims to segment objects without any annotations. Under conditions of multi-camera inputs, the structural, textural and geometrical consistency among each view can be leveraged to achieve fine-grained object segmentation. To make better use of the above information, we propose Surface representation based Self-supervised Object Segmentation (Surface-SOS), a new framework to segment objects for each view by 3D surface representation from multi-view images of a scene. To model high-quality geometry surfaces for complex scenes, we design a novel scene representation scheme, which decomposes the scene into two complementary neural representation modules respectively with a Signed Distance Function (SDF). Moreover, Surface-SOS is able to refine single-view segmentation with multi-view unlabeled images, by introducing coarse segmentation masks as additional input. To the best of our knowledge, Surface-SOS is the first self-supervised approach that leverages neural surface representation to break the dependence on large amounts of annotated data and strong constraints. These constraints typically involve observing target objects against a static background or relying on temporal supervision in videos. Extensive experiments on standard benchmarks including LLFF, CO3D, BlendedMVS, TUM and several real-world scenes show that Surface-SOS always yields finer object masks than its NeRF-based counterparts and surpasses supervised single-view baselines remarkably.
Code is available at: \url{https://github.com/zhengxyun/Surface-SOS}.

\end{abstract}

\begin{IEEEkeywords}
Self-supervised learning, neural surface representation, multi-view object segmentation. 
\end{IEEEkeywords}

\section{Introduction}
\IEEEPARstart{G}{iven} a set of multi-view images or a casually-captured video, segmentation of foreground objects is an important problem in computer vision, with downstream applications in segmentation \cite{view_cons, MATNet} and beyond, such as in video editing \cite{stnerf, controlnerf}, 3D scene understanding \cite{obj_com, ivcc}. Robust object segmentation can now be achieved reliably in scenarios for which large amounts of annotated data are available \cite{maskrcnn, fine_sem, rvm}. However, for less common activities, such as concerts and stage shows, it remains challenging, due to the difficulty in accessing corresponding annotated datasets. Despite some self-supervised approaches \cite{omni, retiming} promising to address this problem, most of them depend on strong constraints, such as the target objects being seen against a static background, or relying on temporal supervision on video. This may result in blurry segmentation masks and false detection of segmentation boundaries. 

\begin{figure}[]
\centering
\includegraphics[width=1.02\linewidth]{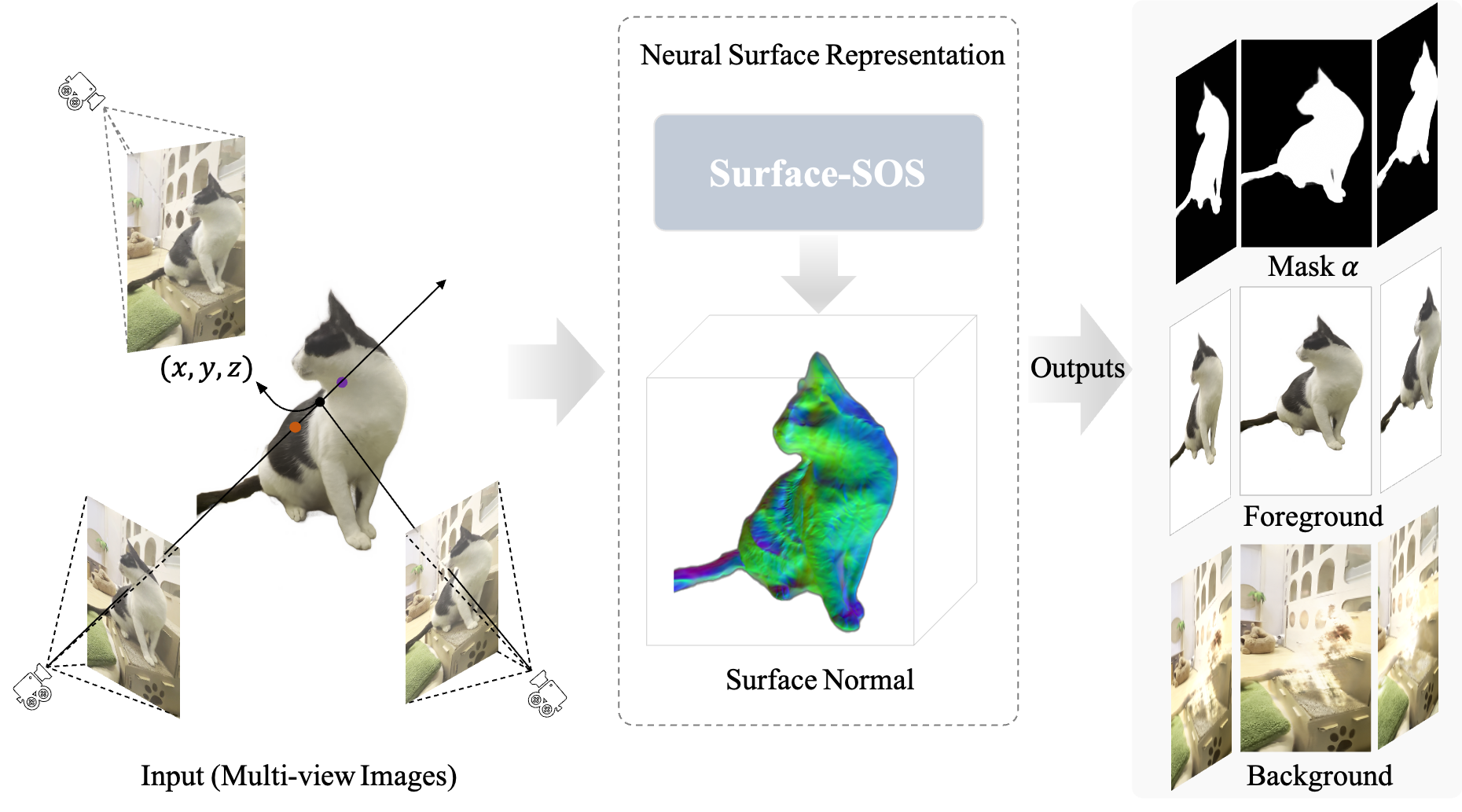} 
\vspace{-6mm}
\caption{We present Surface-SOS, in which multi-view geometric constraints are embedded in the form of dense one-to-one mapping in 3D surface representation. Given multi-view images as input, Surface-SOS predicts convincing results including object masks, foregrounds and backgrounds.}
\label{mvos}
\end{figure}

2D images are the projections of the underlying 3D scenes. Consequently, omitting 3D information may lead to ambiguities in the task resulting from partial occlusion and background confusion, as in most single-view-based approaches. Using several cameras complicates data acquisition but only to a limited extent, as calibrations and predictable setups are available for most applications or can be computed using off-the-shelf tools such as Structure from Motion (SfM) \cite{sfm}. This is also true for the emerging trend of hand-held cell phone, and static camera setups such as performance capture studios \cite{layerseg, Proxemic}. These cameras are readily available, and allow for impromptu shots, as well as quick coverage of large spaces. Under conditions of multi-camera inputs, the structural, textural and geometrical consistency among each view can be leveraged to achieve fine-grained object segmentation \cite{view_cons}. Motivated by the above issues, we reconsider the task of self-supervised segmentation from a 3D perspective, given only 2D images of a scene from multiple viewpoints, the cross-view geometric constraints are embedded in the form of one-to-one dense mapping in 3D space, see Fig. \ref{mvos} for example. This is an intrinsically challenging problem, especially when the number of views is small, or viewpoints far apart. 

The emerging neural implicit representation approaches provide promising results in novel view synthesis \cite{nerf, nsvf, instant_ngp} and high-quality 3D reconstruction from multi-view images \cite{sdfdiff, neus, neus2}. The neural volume rendering approach presented in \cite{nerf} and the follow-up works \cite{semanticNeRF, fastnerf} have recently shown that representing both the density and radiance fields as neural networks can lead to promising novel view synthesis results from a sparse set of images. Such approaches significantly transfer multi-view information between views without explicit reconstruction of 3D geometry, neural radiance field (NeRF) \cite{nerf} and its many scene-specific NeRF works will likely make a long-lasting impact on semantic scene understanding. Although this coupling indeed leads to a good generalization of novel viewing, the density is not as successful in faithfully predicting the actual geometry of the scene, often producing noisy, low-fidelity geometry approximation. These methods require precise object masks and appropriate weight initialization due to the difficulty of propagating gradients \cite{semanticNeRF}. Due to this sampling imbalance, volumes that are close to the camera receive significantly more gradients \cite{floaters}. This can lead to incorrect density buildup and result in floating artifacts. Moreover, it is generally difficult when the object of interest is severely occluded, with weak texture or with a similar appearance to the background. The seminal neural implicit surface representation \cite{neus, VolSDF} substitutes the density field with a zero-level set of Signed Distance Function (SDF) in the volume rendering formulation, leading to a better approximation of the geometry while maintaining the quality of view synthesis and high-quality 3D reconstruction.

In light of this, we take one step further to investigate how to effectively segment objects from a 3D perspective. We present a framework, named Surface Representation for Self-Supervised Object Segmentation (Surface-SOS), connects object segmentation and neural surface representation to segment objects within a complex scene.  We propose a neural scene decomposition scheme that decouples the 3D scene into two complementary neural representation modules for both the foreground and background: a Foreground Consistent Representation (FoCoR) module and a Background Completion (BaCo) module. By connecting the SDF-based surface representation to geometric consistency, and applying volume rendering to train this representation with robustness, we can reconstruct the foreground object geometry and background appearance from multi-view images. To accelerate the training process, and apply volume rendering to train this representation with robustness, we incorporate SDF volumetric rendering with multi-resolution hash encodings \cite{instant_ngp}.
Moreover, we propose several critical training strategies for faster training convergence and better surface representation, favoring better foreground and background decomposition with fine detail.

The proposed method takes a sequence of multi-view images as input, and estimates a dense, geometrical consistent object map, as well as a textural, completed background for each view. Such a representation characterizes the compositional nature of scenes and provides additional inherent information, thus benefiting 3D scene decomposition. We validate the effectiveness of Surface-SOS using multi-view datasets and monocular stereo video, including public benchmarks for real-world forward-facing datasets (LLFF \cite{llff}), object-centric datasets BlendedMVS \cite{blendedmvs}, Common Objects in 3D (CO3D \cite{co3d}), and TUM dataset \cite{tum}; and real-world RGB video captures from \cite{consistent}. Extensive experiments show that Surface-SOS performs better than the state-of-the-art (SOTA) image-based object (co-) segmentation frameworks and NeRF-based object segmentation methods. Even without auxiliary inputs such as object mask, Surface-SOS can effectively recover dense 3D surface structures from multi-view images that lead to photo-realistic rendering results and high-quality segmentation maps. 

In summary, this paper makes the following contributions:
\begin{itemize}
\item We present Surface-SOS,  a new self-supervised approach that leverages neural surface representation to break the dependence on large amounts of annotated data and strong constraints to achieve superior performance. 
\item We design a 3D scene decomposition scheme containing two complementary neural representation modules for both foreground and background. By leveraging the multi-resolution hash grid SDF, Surface-SOS can generate compact geometric surfaces, and produces finer object segmentation compared to its NeRF-based counterparts.  
\item We introduce several critical strategies for faster convergence and better surface representations. Our proposed framework can be implemented on general benchmarks for forward-facing, object-centric, indoor, and real-world dynamic scenarios. 
\item Extensive experiments and ablation studies justify the design of each component and demonstrate that Surface-SOS yields finer object segmentation than its NeRF-based counterparts. It also remarkably improves single-view methods by simply adding original masks as an additional input and generating fine-grained segmentation.
\end{itemize}

\begin{figure*}[]
\centering
\includegraphics[width=\linewidth]{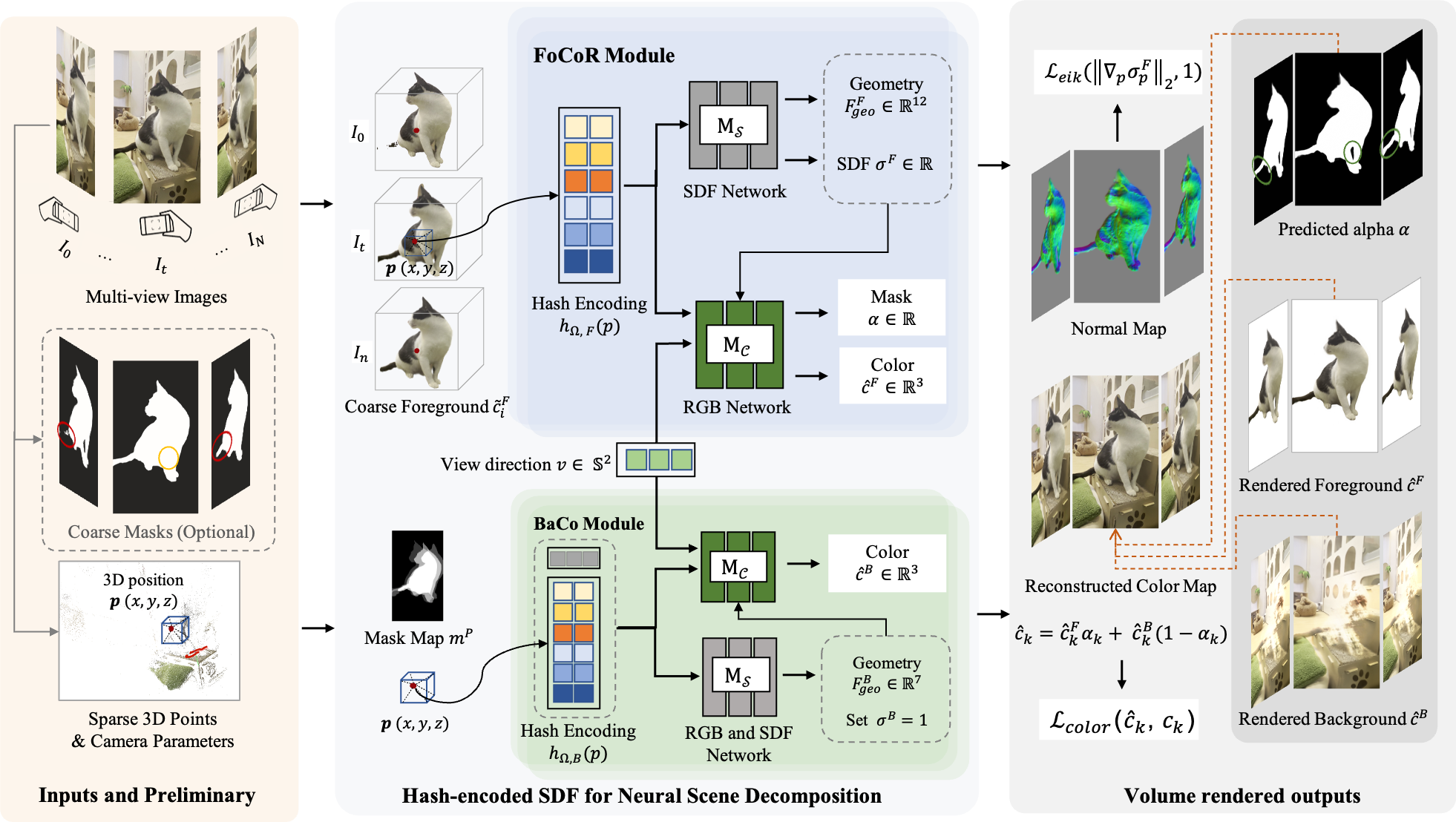} 
\vspace{-6mm}
\caption{{\bf{Method overview}}. For the scene captured by N images $\{{I_i}\}^N_{i=1}$, we use COLMAP \cite{colmap} and Mask-RCNN \cite{maskrcnn} to get sparse 3D points and coarse object masks as co-inputs, and predict a dense, geometrical consistent object map, as well as a textural, completed background for each image. Note that the coarse mask is optional and merely expedites the convergence of 3D surface representation. Moreover, by introducing coarse masks as additional input, Surface-SOS is able to refine segmentation remarkably (see the under-segmentation and over-segmentation highlighted in \textcolor{Maroon}{red} and \textcolor{Goldenrod}{yellow}, respectively). Surface-SOS consists of two complementary representation modules: a Foreground Consistent Representation (FoCoR) module and a Background Completion (BaCo) module. {\bf{FoCoR}}: For every image, given a 3D point $p(x,y,z)$, we concatenate its queried feature from the multi-resolution hash grid as the input to the SDF network. The SDF network outputs the geometry feature and SDF value, which are combined with the viewing direction and further fed into the RGB network to generate RGB value for the foreground, as well as the alpha $\alpha$ prediction. {\bf{BaCo}}: Given a sequence of multi-view images, we concatenate its static features from the multi-resolution hash grid and its 3D position $p(x,y,z)$ as the input to the SDF network. Here, we crop the foreground from the probability map region $m^P$ by setting the SDF value to a positive number (e.g. 1.0). Then the SDF value $\sigma^B$ and geometry feature vectors $\mathbf{F}_{geo}^B$ are combined with the viewing direction $\mathbf {v} \in \mathbb{S}^2$ and further fed into the RGB network $\mathrm{M}_{\mathcal{C}} $ to generate the RGB value for the background $c^B$. After removing the foreground from the probability map $m^P$, even though some parts of the background were occluded in the original view, the other views of the scene provide sufficient textural/structural information to complete the missing background. All parts of the proposed pipeline are trained end-to-end with the geometric and photometric losses in a self-supervised manner with the original input images.}
\label{framework}
\end{figure*}

\section{Related Work}

\subsection{Single View Object Segmentation}
Object segmentation is a longstanding problem in computer vision. Most object segmentation algorithms are fully supervised \cite{maskrcnn, pointrend} and require large annotated datasets containing pairs of images and labels \cite{segany}. Our goal is to train a purely self-supervised method without either segmentation or object bounding box annotations. Many approaches take advantage of the motion patterns of objects as complementary cues \cite{motion, cmsalgan}, which use a two-stream network to process the RGB image and the corresponding optical flow separately and fuse the results in the end. To avoid the expensive computation of optical flow, some work \cite{target, uovos} utilizes higher-order spatial and temporal relations between video frames to bring more comprehensive content understanding. However, these motion-based segmentation methods are prone to accumulate errors calling for a new system with high accuracy performance and robustness on each frame. Furthermore, motion-based object segmentation relies on masks or uses task-specific training datasets, which lack the capability of preserving fine details, e.g., human hairs and animal fur.

Image matting deals with the problem of estimating an RGBA foreground (color image + alpha matte) and a background color image from a given image \cite{matte, himm, trimap_matte}. Mathematically, an image $I$ can be viewed as the linear combination of a foreground $F$ and a background $B$ through an $\alpha$ coefﬁcient: $ I = \alpha F + (1-\alpha) B $. With the help of trimaps, image matting predicts a detailed alpha matte which can be used to recover the mixing factor of foreground and background \cite{poisson_matte, trimap_matte}. Extensive research has provided promising performance on deep learning-based video matting \cite{bgmv2, rvm}. However, existing supervised deep models require enormous manual annotations for training. Furthermore, if the training data do not adequately cover the sampling variation, the trained model may be biased and may not generalize well for images that do not correlate strongly with the training data. It still in most cases generated and augmented the training samples by compositing the images with various background images and foreground, while it makes the synthesized images into unreal scenarios, and produces unacceptable alpha matte. 

The recent video layered representations leverage the power of deep neural networks to separate a moving object in a video from its background by representing a video as a composition of layers with simpler motions \cite{retiming, omni}. They use neural rendering and fit layer models to images by optimizing transformations to minimize reconstruction loss. However, with a large number of layer decompositions that could completely reconstruct the video while outputting nonsensical group separations, this is a difficult problem to solve. The Layered Neural Atlases (LNA) \cite{lna} adopts multi-layer perceptrons (MLPs) to decompose and map the video into sets of 2D atlases. Consequently, MLP-based atlases lead to better decomposition than a standard fixed pixel grid atlas, owing to the image representations being continuous with respect to spatial or spatio-temporal pixel coordinates. As mentioned above, the ideas of layer decomposition have demonstrated the effectiveness of motion as input or supervision for segmentation. However, motion signals can be uninformative or even dishonest in cases such as deformable objects and objects with reflections or occlusions, resulting in unacceptable segmentation.

\subsection{Co-segmentation Approaches}
Co-segmentation is the task of detecting and segmenting the common objects from an image pair and by extension to more images \cite{CycleSegNet}. The key assumptions of these methods are the observation of a common foreground region, or objects with the same appearance characteristics, as opposed to a background with higher variability across images. Initially, a deep dense conditional random field \cite{densecoseg} is applied to the co-segmentation task. They use a co-occurrence map to measure the objectness for object proposals, and the similarity evidence for proposals is generated by a selective search which uses scale-invariant feature transform (SIFT) feature. An end-to-end training framework \cite{deepcoseg} introduces convolutional neural network (CNN) to jointly detect and segment the common object from a pair of images. A later attention-based method \cite{semantic_coseg} takes the encoded features to pay attention to the common objects via a semantic attention learner. Most recently, a new co-segmentation framework \cite{dino_coseg} based on the deep features extracted from a pre-trained Vision Transformer (ViT) has been proposed and achieves better results on object co-segmentation and part co-segmentation.
 
\subsection{Multi-View Object Segmentation}
The above approaches to object segmentation usually require the user to provide information about background or foreground in advance. To avoid ambiguity between foreground and background model, many researchers have attempted to solve this problem automatically by using additional information such as stereo cues. Early attempts \cite{3dseg, rang_color} to segment an object from multiple views by combining photometric information with depth information from stereo images. These approaches provide much better results than methods relying solely on color information, but they are designed for stereo imaging systems and cannot be readily applied to systems with more than two cameras. Some approaches for multi-view segmentation transfer information between views without explicit representations of 3D geometry \cite{silhouette_extra, view_cons}. The first attempt to solve the multi-view segmentation problem uses a silhouette-based algorithm \cite{silhouette_extra}. A later method \cite{silhouetteseg} focuses on probabilistic occupancy along viewing lines. An inter-view consistent approach links superpixels between images \cite{view_cons}, where geometric cues are propagated using camera parameters to ensure consistency between views. This approach adopts the assumption that color values of a foreground object are different from those of the background region. Indeed, it then becomes difficult to rely on shared appearance models of the object between views while parts of the background seen from several viewpoints will present similar aspects. In contrast, our approach aims towards delicate segmentation within a complex scene, by combining the neural implicit representation power from multi-view images in an end-to-end and self-supervised manner.

\subsection{Neural Implicit Representation}
The neural implicit representation has shown its effectiveness in novel view synthesis \cite{nerf, nsvf, instant_ngp} and high-quality 3D reconstruction from multi-view images \cite{sdfdiff, neus, neus2}. The neural implicit exploit the coordinate-based representation to model the scene by querying points attributes with their 5D coordinates $ (x, y, z, \theta, \phi)$. Such representation significantly improves traditional image-based modeling or rendering in an end-to-end and self-supervised manner. NeRF \cite{nerf} and its many scene-specific NeRF works will likely make a long-lasting impact on semantic scene understanding. In particular, Semantic-NeRF \cite{semanticNeRF} adds an extra head to NeRF to predict semantic labels at any 3D position. As a supervised approach, Semantic-NeRF requires semantic labels for full supervision. The recent work NeRF-SOS \cite{nerfsos} presents a framework for learning object segmentation in complex real-world scenes by using a collaborative contrastive loss at the appearance segmentation and geometry segmentation levels. However, the optimization process for training is affected by the conflicting update directions, contrastive loss, and ViT semantic feature extraction, which remains a notorious challenge in multi-task learning. RFP \cite{rfp} is an unsupervised multi-view image segmentation using radiance field propagation with a bidirectional photometric loss to guide the reconstruction of semantic field. This approach adopts the assumption that there is no object motion in the scene, or color values of a foreground object are different from those of the background region. However, given only a set of calibrated input images, the NeRF reconstruction problem is ill-posed. NeRF acquisition typically suffers from background collapse, creating near-camera floating artifacts on the edges of the captured scene. Several works \cite{mip360, floaters} observed and proposed solutions to the problem of floaters and background collapse. The main insight is that the floating artifacts occur due to a higher density of samples in regions near cameras \cite{floaters}. Similarly, without visual cues, the model has lost the perception of the object. Nevertheless, extending the semantics and discovery of object decompositions with NeRF is not trivial, as they cannot extract high-quality surfaces since the geometry representation does not contain surface constraints. The techniques are generally difficult when the object of interest is highly occluded, has a weak texture, or has a similar appearance to the background. 

Early work focused on predicting the geometry of shapes using occupancy fields \cite{occupancy} or SDF \cite{deepsdf}. NeuS \cite{neus} represents the 3D surface as an SDF for high-quality geometry reconstruction. This allows for differentiable rendering by tracing rays through the scene and integrating over them. However, the explicit integration also makes this approach very computationally intensive, training NeuS is very slow, and it only works for static scene reconstruction. To overcome the slow training time of deep coordinate-based MLPs, Instant-NGP \cite{instant_ngp} proposes a multi-resolution hash encoding and proves its effectiveness to speed up. With this in mind, we present a self-supervised learning framework for multi-view object segmentation by leveraging the geometric consistency of instant neural surface representation. Unlike previous works that assume a static camera or background, we allow the representation network to deform space along with the parallax or pose changes, then reconstruct the geometry and appearance of the foreground, as well as a textured, complete background.

\section{Proposed Method}
\label{Method}

\subsection{Overview}
Surface-SOS takes a sequence of multi-view images as input, and estimates a dense, geometrical consistent object segmentation map, as well as a textural, completed background for each view. Fig. \ref{framework} shows an overview of our method. To tackle this challenging task by leveraging the existence of geometric consistency of the one-to-one dense mapping in 3D space, we decouple the scene into two complementary neural scene representation modules: a Foreground Consistent Representation (\textit{FoCoR}) module and a Background Completion (\textit{BaCo}) module. We build our scene representation modules upon the SDF-based neural surface representation, and incorporate multi-resolution hash encodings for training acceleration. We introduce geometric and photometric losses to train the network in a self-supervised manner. Moreover, we propose several critical training strategies for faster training convergence and better surface representation, which favors better foreground and background decomposition with fine detail. Given a 3D spatial point, we concatenate its queried feature from the multi-resolution hash grid and its 3D position as input to the SDF-based networks. The outputs of SDF network are combined with the viewing direction and further fed into the RGB network. For FoCoR, the RGB network predicts color and alpha values for the foreground, while for BaCo, the network predicts the background color.

\subsection{Preliminary}

{\bf{NeuS}}. Given a set of posed multi-view images, the scene of the object is represented by two functions: a signed distance field $f(\mathbf {p}): \mathbb{R}^3 \to \mathbb{R}$ that maps a spatial position $\mathbf{p} \in \mathbb{R}^3$ to its signed distance to the object, and a radiance field $c(\mathbf {p}, \mathbf {v}): \mathbb{R}^3 \times \mathbb{S}^2 \to \mathbb{R}^3$ that encodes the color associated with a point $\mathbf{p} \in \mathbb{R}^3$ and a view direction $\mathbf {v} \in \mathbb{S}^2$. The surface $\mathcal{S}$ of the object can be obtained by extracting the zero-level set of the SDF $\mathcal{S}=\left\{\mathbf{p} \in \mathbb{R}^3 \mid f(\mathbf{p})=0\right\}$. Then the object is rendered into an image by volume rendering. Specifically, for each pixel of an image, we sample $n$ points $ \{{p(t_i) = \mathbf {o} + t_iv|i = 0, 1, . . . , n - 1} \}$ along its camera ray, where $\mathbf{o}$ is the center of the camera and $\mathbf {v}$ is the view direction. By accumulating the SDF-based densities and colors of the sample points, we can compute the color $\hat{C}$ of the ray. As the rendering process is differentiable, NeuS can learn the signed distance field $f(\mathbf {p})$ and the radiance field $c(\mathbf {p}, \mathbf {v})$ from the multi-view images. However, the training process of NeuS is very slow, taking about 8 hours on a single GPU.

{\bf{Multi-resolution Hash Encoding}}. To overcome the slow training issue of deep coordinate-based MLPs (which is also a major issue for slow performance of NeuS), multi-resolution hash encoding was proposed \cite{instant_ngp} and shown to be effective. Specifically, it assumes that the object to be reconstructed is bounded in multi-resolution voxel grids. The voxel grids at each resolution are mapped to a hash table with a fixed-size array of learnable feature vectors. For a 3D position $\mathbf{p} \in \mathbb{R}^3$, it obtains a hash encoding at each level $h^i(p) \in \mathbb{R}^d$ ($d$ is the dimension of a feature vector, $ i = 1, ..., L$) by interpolating the feature vectors assigned at the surrounding voxel grids. The hash encodings at all $L$ levels are concatenated into $h(p) = \{{h^i(p)}\}^L_{i=1} \in \mathbb{R}^{L \times d}$ to be the multi-resolution hash encoding.

\subsection{Neural Scene Decomposition via Hash-encoded SDF}
\textbf{FoCoR Module}. For every image, given a 3D position $\mathbf{p}(x,y,z) \in \mathbb{R}^3$ in the rough foreground object, we map its multi-resolution hash encodings $h_{\omega}(p) \in \mathbb{R}^{L \times d}$ with learnable hash table entries $\omega$. We concatenate its acquired feature vectors from the multi-resolution hash grid and the 3D position as the input to the SDF network $ \mathrm{M}_{\mathcal{S}}$, which consists of a shallow MLP: 

\begin{equation}
 (\sigma ^ {F}, \mathbf{F}_{geo}^F)= \mathrm{M}_{\mathcal{S}} \left(\mathbf{p}, h_{\Omega, F}(\mathbf{p})\right).
\end{equation}

The SDF network outputs the SDF value $\sigma^F \in \mathbb{R}$ and geometry feature vectors $\mathbf{F}_{geo}^F \in \mathbb{R}^{12}$, which are combined with the viewing direction $\mathbf{v} \in \mathbb{S}^2$ and further fed into RGB network $\mathrm{M}_{\mathcal{C}}$ to generate RGB value for the foreground object. 

The normal $\mathbf{n}$ of the point $\mathbf{p}$ can be computed as $\mathbf{n}=\nabla_{\mathbf{p}} \sigma^F$ by the gradient of the SDF. We observe that the RGB network is biased to output similar colors for neighboring sample points when their corresponding surface normal is close. Adding normals to the input encourages the reconstructed surface to be smoother, especially for texture-less areas. Eventually, we feed the normal $\mathbf{n}$ with the SDF value $\sigma ^F$, the geometry feature $\mathbf{F}_{geo}^F$, the point $\mathbf{p}$, and the ray direction $\mathbf {v}$ to the RGB network $\mathrm{M}_{\mathcal{C}}$, then the foreground appearance and alpha prediction is formulated as

\begin{equation}
 c^F=\mathrm{M}_{\mathcal{C}} (\mathbf{p},\mathbf{n},\mathbf{v},\sigma^F,\mathbf{F}_{geo}^F).
\end{equation}

\begin{equation}
 \alpha =\mathrm{M}_{\mathcal{C}}  (\mathbf{p},\mathbf{n},\mathbf{v},\sigma^F,\mathbf{F}_{geo}^F).
\end{equation}

{\bf{BaCo Module}}. Training neural surface representation network without a background model can lead to floaters (uncontrolled object surfaces) in free space \cite{neus}. One of the main reasons for this problem is the floaters in the background color, which do not affect the rendering quality and therefore cannot be optimized when training with only photometric loss. Since we want to segment and decouple the foreground and background components of the scene, we circumvent this problem by applying a static background module $c(\mathbf {p}, \mathbf {v}): \mathbb{R}^3 \times \mathbb{S}^2 \to \mathbb{R}^3$ to inpaint static background from the cropped mask region $M_p$. The mask $m^P \in [0, 1]$ indicates the likelihood that a point belongs to the swept volume of the foreground in the scene. The SDF network outputs the SDF value $\sigma^B \in \mathbb{R}$ and geometry feature vector $\mathbf{F}_{geo}^B \in \mathbb{R}^{7}$. Here, we crop the foreground from the probability map region $m^P$ by simply setting the SDF value to a positive number (e.g. 1.0). Since we assume a unimodal (i.e. bell-shaped) density distribution centered at 0 \cite{neus}, an area with a large SDF value is less likely to be sampled, and therefore appears transparent. Then the density value $\sigma^B$ and the geometry feature vector $\mathbf{F}_{geo}^B$ are combined with the viewing direction $\mathbf {v} \in \mathbb{S}^2$, to be further fed into the RGB network $\mathrm{M}_{\mathcal{C}} $ to generate RGB value for the background $c^B$. After removing the foreground from the probability map $m^P$, even with parts of the background occluded in the original view, the other views of the scene provide sufficient textural/structural information to complete the missing background.

\begin{equation}
 c^B=\mathrm{M}_{\mathcal{C}}  (\mathbf{p},\mathbf{n},\mathbf{v},m^P, \sigma^F,\mathbf{B}_{geo}^B).
\end{equation}

Finally, the RGB color of a 3D position $\mathbf{p}$ can be reconstructed by alpha-blending the corresponding points:

\begin{equation}
 c= \alpha c^F + (1-\alpha)c^B.
\end{equation}

\begin{figure}[]
\centering
\includegraphics[width=0.5\textwidth]{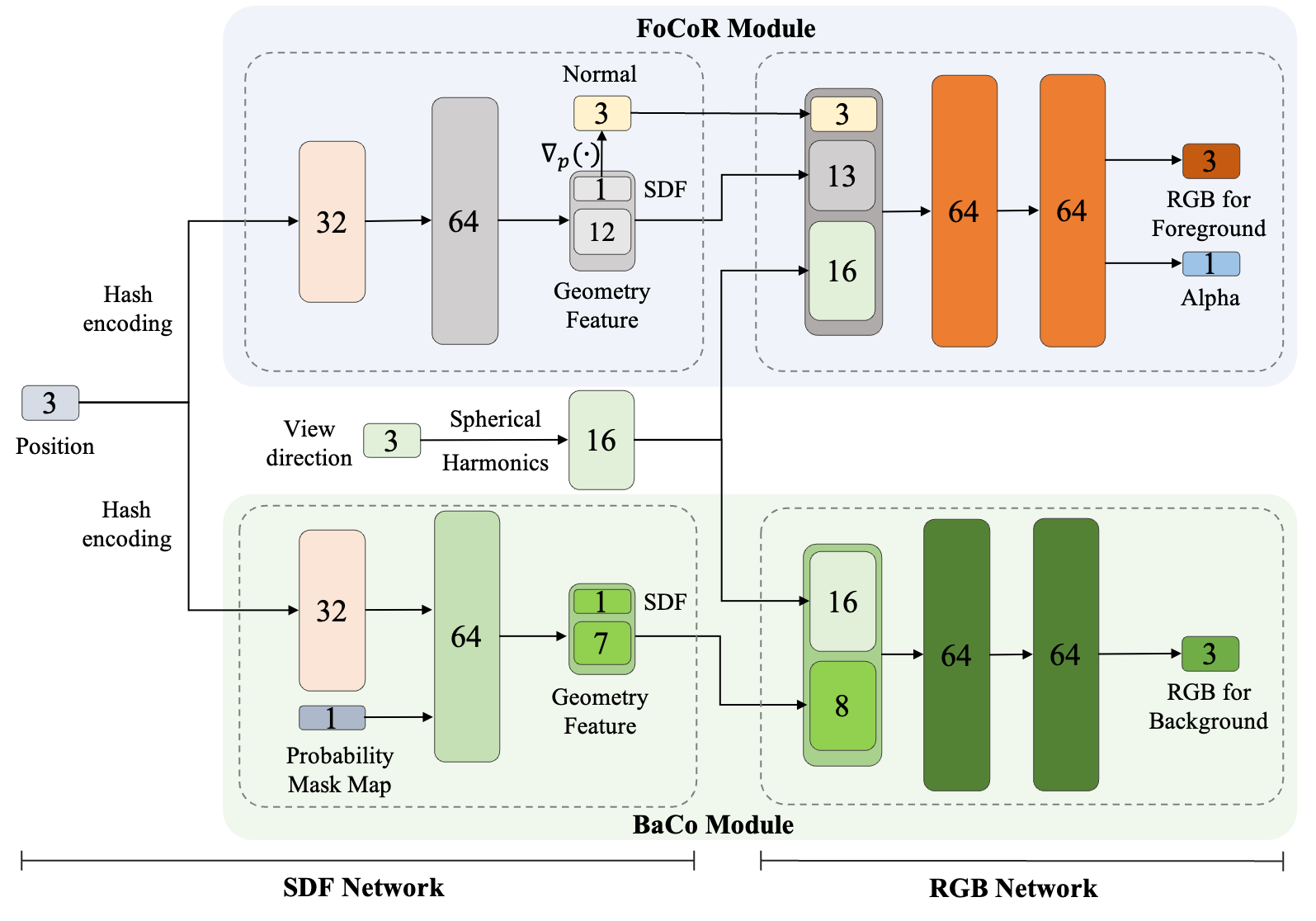} 
\vspace{-6mm}
\caption{A visualization of the architecture of FoCoR and BaCo module.}
\label{architecture}
\vspace{-6mm}
\end{figure}

As shown in Fig. \ref{architecture}, the network architecture consists of the following components: (a) two multi-resolution hash grids with 16 levels of different resolutions ranging from 16 to 2048; (b) two SDF networks modeled by a 1-layer MLP with 64 hidden units; (c) an RGB network modeled by a 2-layer MLP with 64 hidden units for consistent foreground object; (d) an RGB network modeled by a 2-layer MLP with 64 hidden units for static background.

\subsection{Supervision and Volume Rendering Outputs}
{\bf{Volume Rendering}}. To learn the parameters of the neural SDF and color field, we apply the unbiased volume rendering scheme to render images from the Hash-encoded SDF representation. Given a pixel, we sample $n$ points $\{{p(t_i) = \mathbf {o} + t_iv|i= 0,1,...,n}\}$ along its camera ray, where $\mathbf {o}$ is the camera center and $\mathbf {v}$ is the unit direction vector of the ray. 

To obtain discrete counterparts of the opacity and weight function, we still need to adopt an approximation scheme, which is similar to the composite trapezoid quadrature. By using opaque density $\sigma ^ {F}$, the alpha values $\alpha$ are defined in discrete form by

\begin{equation}
\alpha_i=\max \left(1-\frac{\Phi_b\left(f\left(\mathrm{p}\left(t_{i+1}\right)\right)\right)}{\Phi_b\left(f\left(\mathrm{p}\left(t_i\right)\right)\right)}, 0\right),
\end{equation}
where  $\Phi_{b}(x) = 1/(1 + e^{-bx})$ known as the cumulative density distribution, $b$ is a trainable hyperparameter and gradually increases to a large number as the network training converges. 

By accumulating the redefined $\alpha$ values and colors of the sample points, the final color $\hat{c}$ along the ray is computed via the approximation scheme as

\begin{equation}
\hat {c}_(\mathbf{o},\mathbf{v})=\sum_{i=1}^{n}T(t_i)\alpha(t_i)c(\mathbf{p}(t_i), \mathbf{v}),
\end{equation}
where $T(t_i)$ is the discrete accumulated transmittance defined by $T\left(t_i\right)=\prod_{j=0}^{i-1}\left(1-\alpha\left(t_j\right)\right)$, $c(p(t), \mathbf {v})$ means the color at the point $\mathbf {p}$ along the viewing direction $\mathbf {v}$. 

Additionally, we adopt a ray marching acceleration strategy \cite{instant_ngp} to maintain an occupancy grid that roughly marks each voxel grid as empty or non-empty. The occupancy grid can effectively guide the marching process by preventing sampling in empty spaces, and accelerate the volume rendering process.

{\bf{Training and Supervision}}. With the multi-view images as the main supervision signal used to train our system, we introduce photometric and geometric loss to supervise their training. Specifically, we optimize our neural networks and inverse standard deviation by randomly sampling a batch of pixels and their corresponding rays in the world space $P = \{{c_k, M_k, \mathbf {o}_k, \mathbf {v}_k} \}$, $k \in \{{1, ..., m}\}$ from an image in every iteration, where $m$ denotes the batch size, $c_k$ is its pixel color and $M_k \in \{{0, 1}\}$ is its coarse mask value. The final joint loss function is defined as

\begin{equation}
 \mathcal{L} = \mathcal{L}_{\text {color}} + \lambda_e \mathcal{L}_{\text {Eikonal}} + \lambda_s\mathcal{L}_{\text {sparsity}}.
\end{equation}

First, we minimize the distance between rendered pixels $\hat{c}_k$ and the ground truth pixels $c_k$, by using a color loss $\mathcal{L}_{\text {color}}$ defined as

\begin{equation}
\mathcal{L}_{\text {color }}=\frac{1}{m} \sum_k \mathcal{R}\left(\hat{c}_k, c_k\right).
\end{equation}

Here we choose the L2 loss as $\mathcal{R}$, which is shown to be robust to outliers and stable in training.

An important property of SDF is its unit norm. The Eikonal term \cite{eikonal} is therefore added to regularize the learned signed distance field. Specifically, Eikonal loss $\mathcal{L}_{\text {Eikonal}}$ is added on the normal of sampled points:

\begin{equation}
\mathcal{L}_{\text {Eikonal}}=\frac{1}{m n} \sum_{k, i}\left(\left\|\mathbf{n}_{k, i}\right\|-1\right)^2 ,
\end{equation}
where $i$ indexes the $i$ th sample along the ray with $i \in \{{1, ..., n}\}$, and $n$ is the number of sampled points. $\mathbf{n}_{k,i}$ is the normal of a sampled point. 

Additionally, the property of accumulated transmittance in volume rendering means that the invisible query samples behind visible surfaces lack supervision. To address this issue of floaters and background collapse, a sparsity regularization term \cite{sparseneus} is incorporated to generate compact geometric surfaces.

\begin{equation}
\mathcal{L}_{\text {sparsity}}=\frac{1}{m n} \sum_{k, i}exp\left(-\tau·|\sigma^F| \right)^2,
\end{equation}
where $|\sigma^F|$ is the absolute SDF value of sampled point, $\tau$ is a hyperparameter to re-scale the SDF value. This term will encourage the SDF values of the points behind the visible surfaces to be far from 0. When extracting 0-level set SDF to generate mesh, this term can avoid uncontrollable free surfaces. The sparsity loss $\mathcal{L}_{sparsity}$ prevents duplicate representations in the foreground, and further encourages the many-to-one mapping of scene points to the sample points. 

We additionally introduce an optional mask loss $\mathcal{L}_{\text {mask}}$ during the initial train phase. We encourage the predicted alpha values to match a coarse input mask that identifies which regions should decouple the foreground and background scenes. We note that we do not require input masks to be precise, as they are used only for initializing the alpha mapping.  And the majority of training happens without this loss, allowing the network to correct for errors. The alpha loss $\mathcal{L}_{\text {mask}}$ is defined as:

\begin{equation}
\mathcal{L}_{\text {mask}} = BCE(M_k, \alpha_k),
\end{equation}
where $BCE$ is the binary cross entropy loss. 

\section{Experiments}
\label{Experiment}

\begin{figure*}[]
\centering
\vspace{-6mm}
\includegraphics[width=\linewidth]{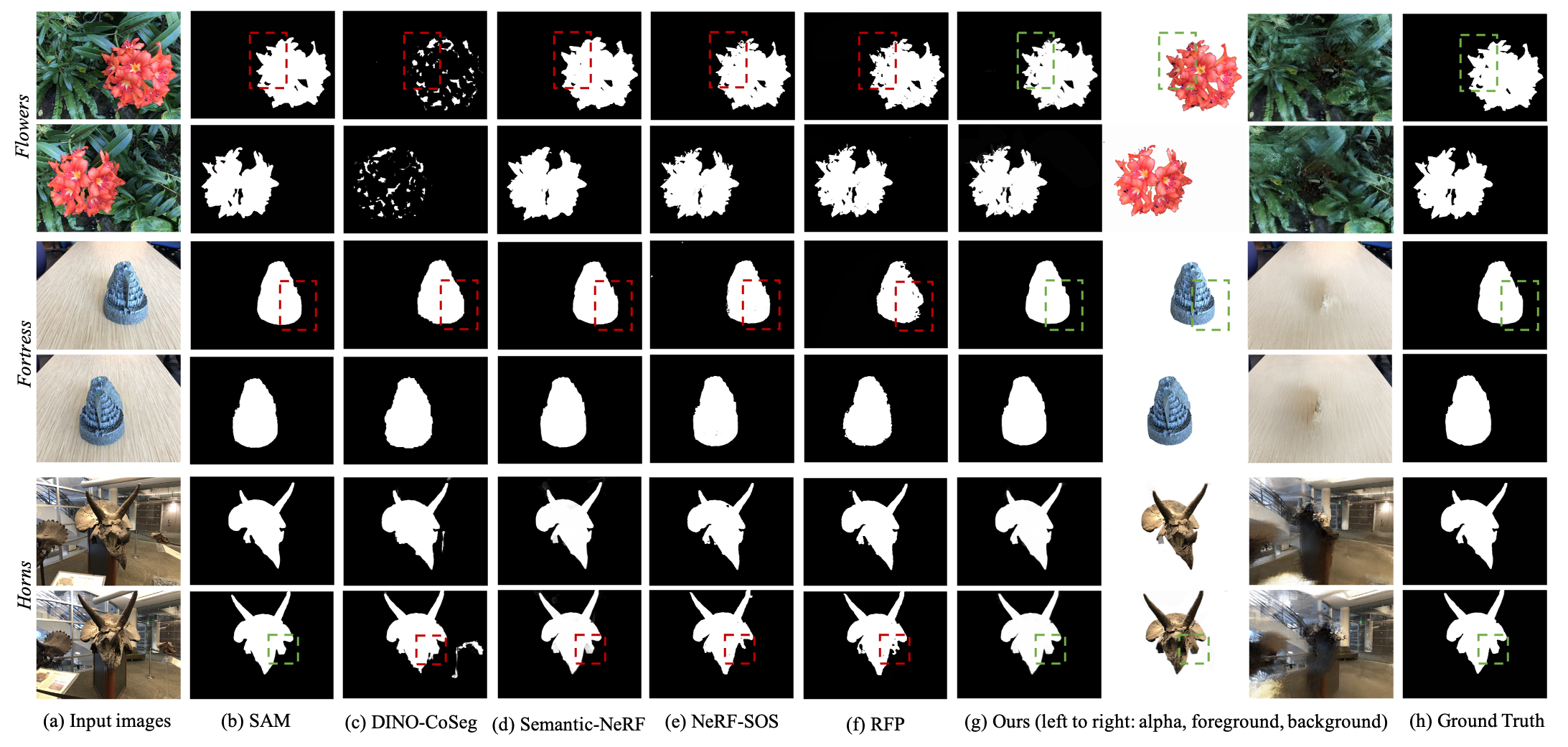} 
\caption{Comparison on the forward-facing scenes \textit{Flower, Fortress}, and \textit{horns} from LLFF dataset \cite{llff}. In the third column, DINO-CoSeg \cite{dino_coseg} mistakenly matches several discrete patches, as DINO has higher activation on just a few tokens, which may lead to view-inconsistent and disconnected co-segmentation results. Compared to SAM \cite{segany} and DINO-CoSeg, our results have more accurate edges, since our network can exploit multi-scale geometry features to better capture the matte objects. Compared with NeRF-based methods (i.e. Semantic-NeRF \cite{semanticNeRF}, NeRF-SOS \cite{nerfsos}, and RFP\cite{rfp}), Surface-SOS (g) produces view-consistent masks with finer details and no holes in the interior of objects. }
\label{cmp_SOTA_llff}
\end{figure*}

\subsection{Experimental Settings and Implementation Details}

\textbf{Datasets and Evaluation Metrics}. We validate the effectiveness of the proposed Surface-SOS on both multi-view benchmark datasets and monocular stereo video data. We provide qualitative and quantitative comparisons with the SOTA object segmentation methods on four publicly available datasets, and choose some representative scenes: 1) LLFF datasets \cite{llff} contain three forward-facing scenes \{\textit{Flower, Fortress, horns}\} with 30 to 62 roughly forward-facing images; 2) BlendedMVS datasets \cite{blendedmvs} contain two object-centric scenes \{\textit{5a6, 5c3}\} with  27 to 110 images; 3) CO3D datasets \cite{co3d} contain two common objects \{\textit{Bicycle, Backpack}\} captured with 100 to 201 images; and 4) two scenes \{\textit{Teddy bear, Plant}\} in the 3D object reconstruction category of the TUM datasets \cite{tum}, sampling with sequences ranging from 175 to 259 frames. We manually labeled all views as faithful binary mask annotations to provide a quantitative comparison for all methods and used them to train Semantic-NeRF \cite{semanticNeRF}. However, these public datasets are either designed for novel view synthesis \cite{llff}, specific domains \cite{blendedmvs}, or videos of static scenes \cite{co3d, tum}. To validate the effectiveness of our approach, we further evaluate on additional more challenging datasets. Specifically, we capture custom stereo video datasets for evaluation, including hand-held phones, and static camera setups such as performance capture studios. Our dataset consists of both static \textit{Dance} and dynamic scenes \textit{Cat} with a gentle amount of object motion. We also leverage three common scenes \{\textit{Kevin, Texting, Boy}\} captured by video sequences from \cite{consistent}. We sample the videos and obtain sequences ranging from 73 to 180 frames.

As for quantitative evaluation, we use the Sum of Absolution Difference (SAD), Mean Square Error (MSE), mean pixel accuracy (Acc.), and mean Intersection over Union (mIoU) as our metrics. Acc. measures the proportion of pixels that have been assigned to the correct region, and mIoU is the ratio between the area of the intersection between the ground-truth segmentation mask and the prediction.

\textbf{Comparison Methods}. Given multi-view images or a casually-captured video, we target to output the corresponding alpha foreground as a segmentation mask. Therefore, several object segmentation baselines are adopted for comparisons: 1) single-view supervised segmentation SAM \cite{segany}; 2) image-based object co-segmentation method DINO-CoSeg \cite{dino_coseg}; and 3) NeRF-based methods, including NeRF-SOS \cite{nerfsos}, RFP\cite{rfp} and supervised Semantic-NeRF \cite{semanticNeRF} trained with annotated masks. On the other hand, video-based foreground matting RVM \cite{rvm}, and the untrained network-based methods LNA \cite{lna} for video layers decomposition are included for a more comprehensive comparison of video sequences. 

\begin{figure*}[]
\centering
\includegraphics[width=\linewidth]{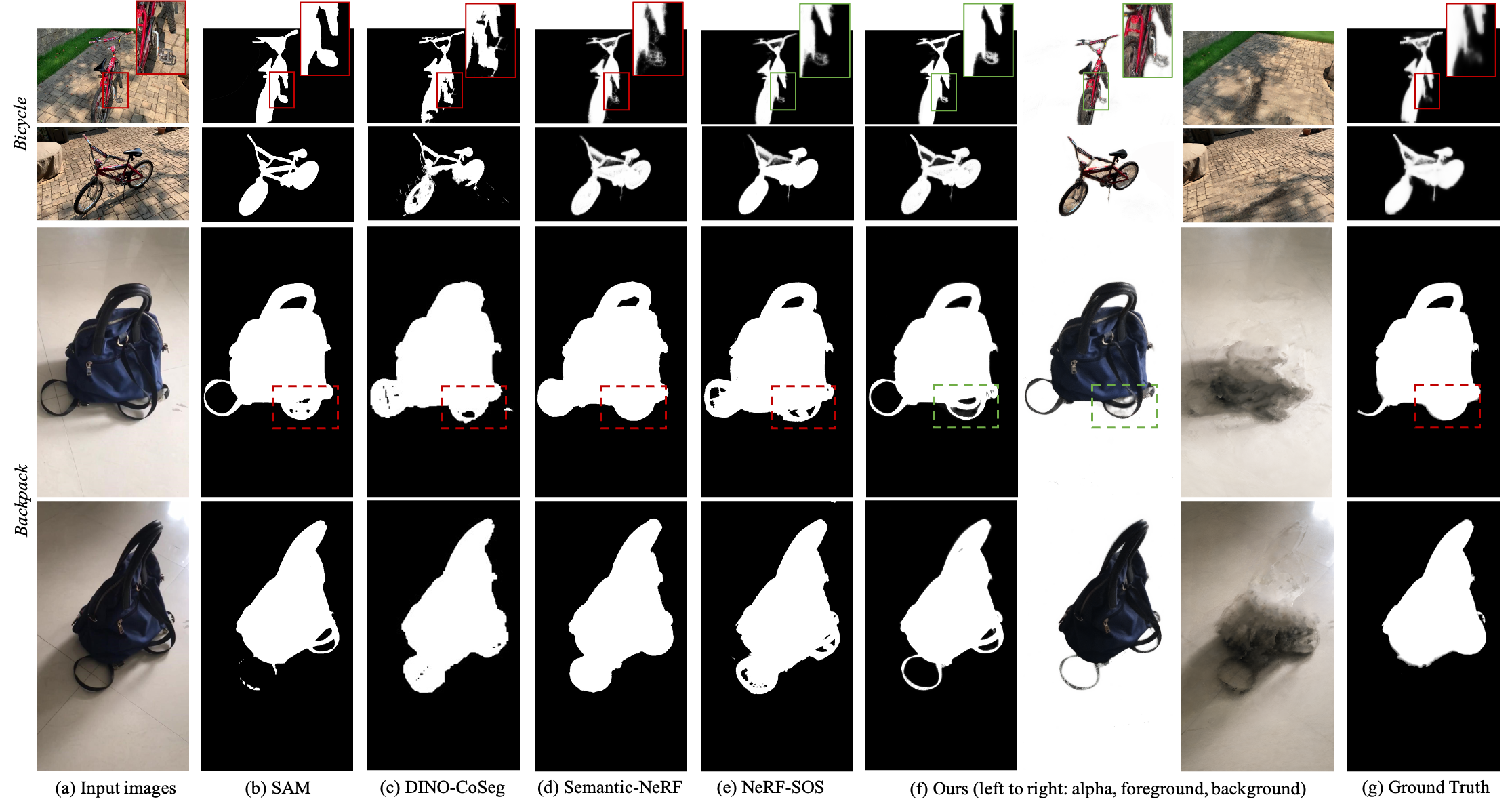} 
\vspace{-6mm}
\caption{Qualitative comparisons on object-centric scenes \textit{Biclcle} and \textit{Backpack} from CO3D data \cite{co3d}. Despite SAM \cite{segany} providing fine-grained boundary information it is noisy and misses more valid detection than ours. Whereas the proposed method achieves high-quality geometric and textural consistent foreground maps without inducing noise, e.g., it can recover the complex structures of the bicycle frame and render detailed textures in the \textit{Bicycle} example. }
\label{cmp_SOTA_co3d}
\end{figure*}

\textbf{Implementation details}. Given a sequence of multi-view images, we first perform an SfM \cite{sfm} reconstruction using an open-source software COLMAP \cite{colmap} to estimate the camera poses and sparse 3D points of the scene. This step provides us with intrinsic and extrinsic camera parameters as well as a sparse point cloud reconstruction. To expedite the convergence of 3D object surface representation, we apply Mask R-CNN \cite{maskrcnn} to segment out the most common foreground in each view independently. SAM \cite{segany} is a general segmentation model trained on a diverse, high-quality dataset of over 1 billion masks, it can produce high-quality object masks from input prompts such as points or object bounding boxes. DINO-CoSeg \cite{dino_coseg} is an image-based object co-segmentation method as it takes a pair of images as input and automatically co-segment semantically common foreground objects. Semantic-NeRF~\cite{semanticNeRF} is a supervised NeRF-based approach as it takes annotated labels as input to supervise a semantic branch for object separation. Thus we feed the ground truth labels as input to these methods where the official implementations are used. NeRF-SOS~\cite{nerfsos} is a self-supervised framework, in which the collaborative contrastive loss is implemented upon the original NeRF \cite{nerf}, and segmentation results are based on K-means clustering. RFP \cite{rfp} is one of the first real-scene NeRF-based approaches for unsupervised multi-view image segmentation. We share the same task of scene object segmentation in 3D perspective with multi-view settings. RFP~\cite{rfp} relies on unsupervised single image segmentation algorithms to get good initialization, thus we provided the initial masks of IEM \cite{iem} as input which was used in RFP~\cite{rfp}. Here we only present the results of RFP on the LLFF dataset, as the official implementation for the other datasets is still not available. Our whole system and all the experiments are implemented on a machine with a single NVIDIA GeForce RTX3090 GPU. We train our models using the ADAM optimizer with a learning rate of 0.01. For each scene representation, we train our model for 40k iterations, which takes around 25 minutes.

\subsection{Qualitative results}

\subsubsection{Comparisons with SOTA on static scenes} For static scene object segmentation, we present examples of the comparison with different baselines on a variety of scenes. CO3D provides ground-truth label maps using PointRend \cite{pointrend}, we create binary masks annotation of other datasets for evaluations and the Semantic-NeRF training with publicly available annotation tool labelme\footnote{https://github.com/wkentaro/labelme}. 

\textbf{LLFF~\cite{llff} Scenes}. As seen in Fig. \ref{cmp_SOTA_llff}, compared to single-view supervised segmentation SAM \cite{segany} and image-based object co-segmentation DINO-CoSeg \cite{dino_coseg}, our method presents more accurate edges, due to the exploitation of multi-scale geometry features to better capture the matte objects. Compared to NeRF-based methods, including NeRF-SOS \cite{nerfsos}, RFP\cite{rfp} and supervised Semantic-NeRF \cite{semanticNeRF}, Surface-SOS produces a more complete foreground matte with no holes in the interior of objects, due to the high-quality geometry representation with the surface constraints.

\begin{figure*}[]
\centering
\includegraphics[width=\linewidth]{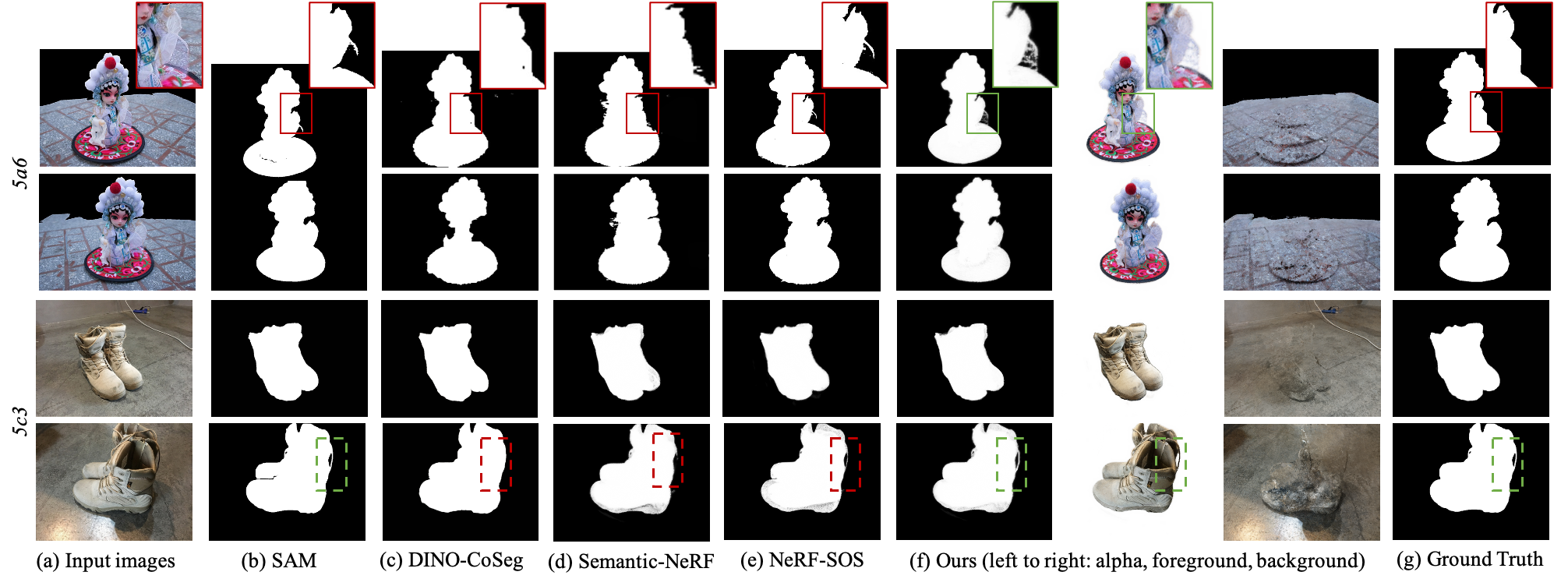} 
\vspace{-6mm}
\caption{Qualitative comparison on the object-centric scenes \textit{5a6} and \textit{5c3} from BlendedMVS dataset \cite{blendedmvs}. Surface-SOS produces more view-consistent masks than other NeRF-based methods. It even generates finer details than the supervised Semantic-NeRF \cite{semanticNeRF} and SAM \cite{segany} (see the openwork sleeve in the top row and the shoelace in the bottom row).}
\label{cmp_SOTA_blender}
\end{figure*}

\begin{figure*}[]
\centering
\includegraphics[width=\linewidth]{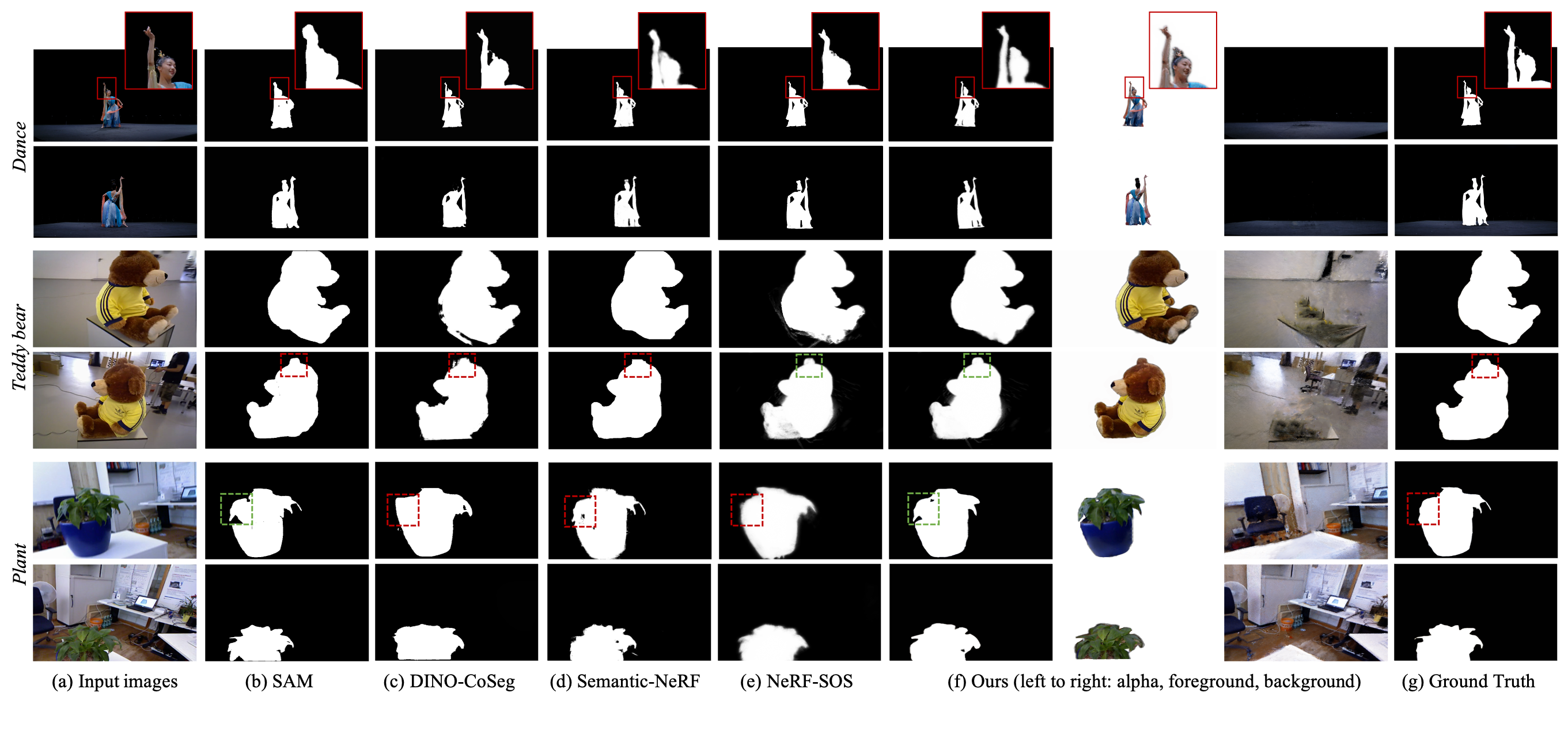} 
\vspace{-6mm}
\caption{Qualitative comparison on the object-centric scenes \textit{Dance, Teddy bear}, and \textit{Plant}. These examples span a wide range of human and non-human species on the complex scene, suggesting the superiority of our proposed methods in the generation of geometrical consistent foreground matte, as well as a textural, completed background. }
\label{cmp_SOTA_static}
\end{figure*}

\textbf{Object-centric Scenes}. Here we use CO3D~\cite{co3d}, BlendedMVS~\cite{blendedmvs}, TUM~\cite{tum} datasets, and the \textit{Dance} scene captured by static camera setups. 

As shown in Fig. \ref{cmp_SOTA_co3d} for the CO3D data, DINO-CoSeg~\cite{dino_coseg} exhibits limited performance in terms of background confusion among foreground objects. The prediction of SAM~\cite{segany} provides fine-grained boundary information but is noisy and lacks the detection of segmentation boundaries. Our method achieves high-quality geometric and textural consistent foreground maps without inducing noise, e.g., it can recover the complex structures of the bicycle frame and render detailed textures in \textit{Bicycle} example scene. 

Fig. \ref{cmp_SOTA_blender} shows comparisons on BlendedMVS dataset \cite{blendedmvs}. From the visualizations, we see that Surface-SOS produces more view-consistent masks than other NeRF-based methods. It even generates finer details than supervised Semantic-NeRF and SAM. For example, in the first row, Surface-SOS can distinguish the openwork details adjacent to the sleeve, and yield accurate segmentation of the shoelace in the bottom row. 

Another comparison of object-centric scenes is shown in Fig. \ref{cmp_SOTA_static}, these examples span a wide range of human and non-human species on the complex scene, demonstrating the superiority of our proposed methods for the geometrical consistent foreground mask, as well as its cross-view correspondence appearance of foreground and background. 

\begin{figure}[htbp]
\centering
\includegraphics[width=\linewidth]{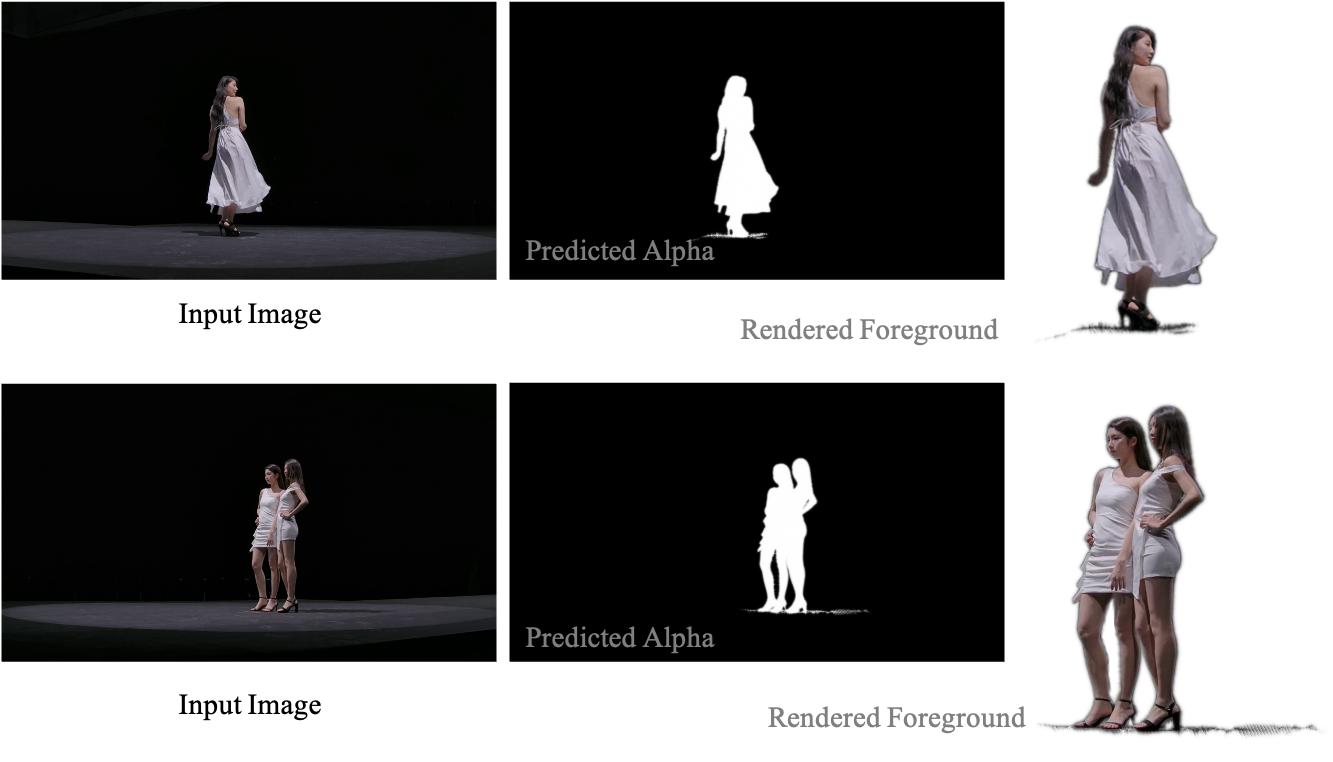} 
\vspace{-6mm}
\caption{More visualizations details of mask and RGB rendering results at the high-resolution of 3840×2160.}
\label{visual}
\vspace{-5mm}
\end{figure}

The more visualizations details of mask and RGB rendering results at the high-resolution of 3840×2160 are shown in Fig.~\ref{visual}. It turns out  that Surface-SOS can produce finer details as the input resolution increases.

\begin{figure*}[htbp]
\centering
\includegraphics[width=\linewidth]{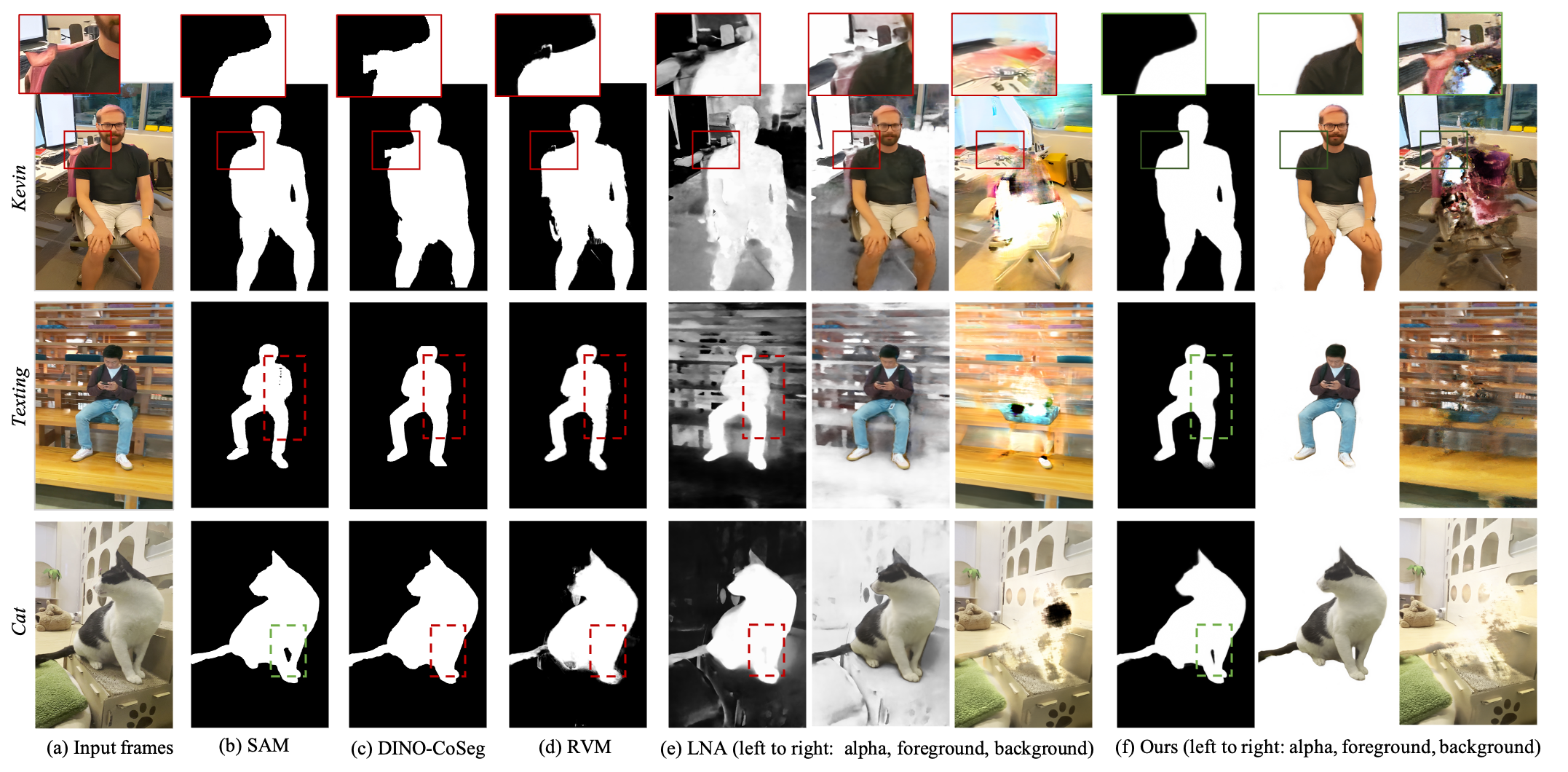} 
\vspace{-6mm}
\caption{Visual comparisons on the dynamic scenes \textit{Texting}, \textit{Kevin} and \textit{Cat}. RVM \cite{rvm} gets excessive smoothing and blurry edges occurring in the high-frequency appearance information on edges and textures. LNA \cite{lna} outputs the nonsensical group separations due to motion signals that may be uninformative or even dishonest in cases such as deformable objects and objects with moderate motion. Our method successfully decomposes temporally and geometrically consistent foreground, as well as textural, complete background.}
\label{cmp_video_lna}
\vspace{-5mm}
\end{figure*}

\subsubsection{Comparisons on dynamic scenes} The above results indicated that Surface-SOS can achieve promising segmentation quality on static scenes. To further evaluate the effectiveness on real scenes, we provide a comparison on our casually-captured videos (i.e., scene \textit{Cat} and the custom stereo video \{\textit{Texting} and \textit{Kevin}\} obtained from \cite{consistent}. Fig.~\ref{cmp_video_lna} presents examples of the qualitative comparison of our method against SAM~\cite{segany}, DINO-CoSeg~\cite{dino_coseg}, RVM \cite{rvm} and the untrained network-based methods LNA \cite{lna} for video layers decomposition. RVM gets excessive smoothing and blurry edges occurred in the high-frequency appearance on edges and textures. LNA outputs the nonsensical group separations due to motion signals that may be uninformative or even dishonest in cases such as deformable objects and objects with moderate motion. As shown in Fig. \ref{cmp_video_lna} (f), our method successfully decomposes the specifying ambiguous foreground and background with two complementary SDF-based representation modules, which is sufficient to obtain visually satisfying results. This suggests that the geometry and appearance cues in forward-backward frames can benefit object segmentation with different viewpoints consistency. Therefore, our method supports video decomposition containing moderate object motion. We encourage readers to review our supplemental videos for a dynamic visualization of qualitative results.

\subsection{Quantitative results}

The quantitative results of compared approaches on four benchmark datasets as well as our captured data (scene \textit{Dance}) are presented in Table \ref{tab:sota}. From the results we can see that, our method outperforms supervised 2D object segmentation methods and the supervised NeRF-based segmentation method (i.e., Semantic-NeRF~\cite{semantic_coseg}). CO3D provides coarse segmentation maps using PointRend \cite{pointrend} while parts of the annotations are missing. Among self-supervised learning frameworks, Surface-SOS performs on par in both evaluation metrics and visualization for view consistency. 

\begin{table}[htbp!]
\caption{Quantitative evaluation of object segmentation on the static scenes. The best results are marked in {\bf Bold Font}. }
\begin{tabular}{lllll}
\toprule[1pt]
Dataset LLFF \cite{llff}                  & SAD $\downarrow$   & MSE $\downarrow$   & mIoU $\uparrow$    & Acc. $\uparrow$  \\ [0.35 ex] 
\midrule
Mask-RCNN \cite{maskrcnn}                 & -                  & -                  & -                  & -              \\  [0.4 ex] 
SAM \cite{segany} (Mask Init.)            & 10.386             & 0.238              & 0.767              & 0.891          \\  [0.4 ex] 
DINO-CoSeg \cite{dino_coseg}              & 9.388              & 0.208              & 0.628              & 0.787          \\  [0.4 ex] 
Semantic-NeRF \cite{semanticNeRF}         & 8.736              & 0.185              & 0.897              & 0.918          \\  [0.4 ex] 
NeRF-SOS \cite{nerfsos}                   & 9.172              & 0.191              & 0.865              & 0.857          \\  [0.4 ex] 
RFP \cite{rfp}                            & 9.494              & 0.229              & 0.782              & 0.829          \\  [0.4 ex]
Surface-SOS (ours)                        & \textbf{8.655}     & \textbf{0.181}     & \textbf{0.903}     & \textbf{0.918} \\  [0.4 ex] 
\midrule
Dataset CO3D \cite{co3d}                  & SAD $\downarrow$   & MSE $\downarrow$   & mIoU $\uparrow$    & Acc. $\uparrow$    \\ [0.35 ex] 
\midrule
Mask-RCNN \cite{maskrcnn} (Mask Init.)    & 4.021              & 0.296              & 0.865              & 0.929          \\  [0.4 ex] 
SAM \cite{segany}                         & 3.265              & 0.226              & 0.876              & 0.940          \\  [0.4 ex] 
DINO-CoSeg \cite{dino_coseg}              & 3.990              & 0.297              & 0.835              & 0.910          \\  [0.4 ex] 
Semantic-NeRF \cite{semanticNeRF}         & 3.534              & 0.272              & 0.851              & 0.924          \\  [0.4 ex] 
NeRF-SOS \cite{nerfsos}                   & 3.588              & 0.275              & 0.844              & 0.915          \\  [0.4 ex] 
Surface-SOS (ours)                        & \textbf{3.011}     & \textbf{0.218}     & \textbf{0.883}     & \textbf{0.945} \\  [0.4 ex] 
\midrule
Dataset BlendedMVS \cite{blendedmvs}      & SAD $\downarrow$   & MSE $\downarrow$   & mIoU $\uparrow$    & Acc. $\uparrow$    \\ [0.35 ex] 
\midrule
Mask-RCNN \cite{maskrcnn} (Mask Init.)    & -                  & -                  & -                  & -              \\  [0.4 ex] 
SAM \cite{segany}                         & 6.963              & 0.165              & 0.929              & 0.936          \\  [0.4 ex] 
DINO-CoSeg \cite{dino_coseg}              & 7.861              & 0.169              & 0.910              & 0.922          \\  [0.4 ex] 
Semantic-NeRF \cite{semanticNeRF}         & 7.915              & 0.191              & \textbf{0.935}     & \textbf{0.955} \\  [0.4 ex] 
NeRF-SOS \cite{nerfsos}                   & 7.739              & 0.192              & 0.924              & 0.934          \\  [0.4 ex] 
Surface-SOS (ours)                        & \textbf{6.872}     & \textbf{0.146}     & 0.931              & 0.941          \\  [0.4 ex] 
\midrule
Dataset TUM \cite{tum}                    & SAD $\downarrow$   & MSE $\downarrow$   & mIoU $\uparrow$    & Acc. $\uparrow$    \\ [0.35 ex] 
\midrule
Mask-RCNN \cite{maskrcnn} (Mask Init.)    & 14.203              & 0.532              & 0.845              & 0.969          \\  [0.4 ex] 
SAM \cite{segany}                         & 13.108              & 0.576              & 0.864              & 0.981          \\  [0.4 ex] 
DINO-CoSeg \cite{dino_coseg}              & 12.949              & 0.546              & 0.821              & 0.966          \\  [0.4 ex] 
Semantic-NeRF \cite{semanticNeRF}         & 13.040              & 0.532              & 0.869              & 0.980          \\  [0.4 ex] 
NeRF-SOS \cite{nerfsos}                   & 12.711              & 0.496              & 0.843              & 0.975          \\  [0.4 ex] 
Surface-SOS (ours)                        & \textbf{9.138}      & \textbf{0.402}     & \textbf{0.870}     & \textbf{0.989} \\  [0.4 ex] 
\midrule
Scene \textit{Dance}                      & SAD $\downarrow$   & MSE $\downarrow$   & mIoU $\uparrow$    & Acc. $\uparrow$    \\ [0.35 ex] 
\midrule
Mask-RCNN \cite{maskrcnn} (Mask Init.)    & 8.066              & 0.490              & 0.726              & 0.820          \\  [0.4 ex] 
SAM \cite{segany}                         & 6.803              & 0.380              & 0.886              & 0.840          \\  [0.4 ex] 
DINO-CoSeg \cite{dino_coseg}              & 6.893              & 0.384              & 0.843              & 0.887          \\  [0.4 ex] 
Semantic-NeRF \cite{semanticNeRF}         & 7.707              & 0.439              & 0.924              & 0.934          \\  [0.4 ex] 
NeRF-SOS \cite{nerfsos}                   & 7.021              & 0.393              & 0.916              & 0.928          \\  [0.4 ex] 
Surface-SOS (ours)                        & \textbf{6.015}     & \textbf{0.355}     & \textbf{0.935}     & \textbf{0.942} \\  [0.4 ex] 

\bottomrule[1pt]
\end{tabular}
\label{tab:sota}
\vspace{-5mm}
\end{table}

\subsection{Ablation Studies}

\begin{table*}[htbp!]
\centering 
\caption{Results in ablation study.}
\begin{tabular}{p{62mm}|llll|llll}
\toprule[1pt]
\multirow{2}{*}{\textbf{}} & \multicolumn{4}{c|}{\textbf{\textit{Teddy}}}            & \multicolumn{4}{c}{\textbf{\textit{Cat}}} \\
         \textbf{} & SAD $\downarrow$ & MSE $\downarrow$ & mIoU $\uparrow$ & Acc. $\uparrow$  & SAD $\downarrow$ & MSE $\downarrow$ & mIoU $\uparrow$ & Acc. $\uparrow$ \\ [0.35 ex] 
\midrule
w/o mask Init. + w/ FoCoR + w/o BaCo               & 16.828   & 0.356  & 0.875   & 0.851                       & 10.603   & 0.586    & 0.761     & 0.724 \\  [0.4 ex] 
w/o mask Init. + w/ FoCoR + w/  BaCo               & 14.188   & 0.517  & 0.857   & 0.899                       & 8.066    & 0.490    & 0.726     & 0.820 \\  [0.4 ex] 
w/ mask Init. + w/o FoCoR + w/o BaCo               & 14.828   & 0.496  & 0.875   & 0.891                       & 8.215    & 0.555    & 0.775     & 0.846 \\  [0.4 ex] 
w/ mask Init. + w/ FoCoR + w/o BaCo                & 11.476   & 0.419  & 0.914   & 0.938                       & 6.892    & 0.383    & 0.871     & 0.887 \\  [0.4 ex] 
w/ mask Init. + w/ FoCoR + w/ BaCo + w/o sparsity  & 9.956    & 0.407  & 0.938   & 0.939                       & 4.192    & 0.383    & 0.926     & 0.961 \\  [0.4 ex]
Surface-SOS (ours)                  & \textbf{8.685} & \textbf{0.328} & \textbf{0.950} & \textbf{0.961}     & \textbf{3.434} & \textbf{0.304} & \textbf{0.946}  & \textbf{0.975} \\  [0.4 ex] 

\bottomrule[1pt]
\end{tabular}
\label{tab:ablation}
\vspace{-5mm}
\end{table*}

\begin{table*}[htbp!]
\renewcommand{\arraystretch}{1}
\centering 
\caption{Quantitative comparison on different single-view object segmentation methods. The best results are marked in {\bf Bold Font}. }
\begin{tabular}{lllll}
\toprule[1pt]
Scene \textit{Texting}                & SAD $\downarrow$                    & MSE $\downarrow$                     & mIoU $\uparrow$                   & Acc.$\uparrow$ \\ [0.35 ex] 
\midrule
Surface-SOS (w/o mask Init.)          & 18.559 (-)                          & 0.386 (-)                            & 0.825 (-)                         & 0.918 (-)              \\ [0.4 ex] 
Mask-RCNN \cite{maskrcnn}             & 19.588                              & 0.390                                & 0.873                             & 0.921                  \\ [0.4 ex] 
Surface-SOS (w/ Mask-RCNN)            & 18.316 ($\downarrow$ 1.272)         & 0.377 ($\downarrow$ 0.013)           & 0.900 ($\uparrow$ 0.027)          & 0.933 ($\uparrow$ 0.012) \\ [0.4 ex] 
SAM \cite{segany}                     & 15.598                              & 0.347                                & 0.955                             & 0.938                     \\ [0.4 ex] 
Surface-SOS (w/ SAM)                  & \textbf{14.828 ($\downarrow$ 0.770)} & \textbf{0.326 ($\downarrow$ 0.021)} & \textbf{0.965 ($\uparrow$ 0.010)} & \textbf{0.951 ($\uparrow$ 0.013) } \\ [0.4 ex] 
RVM \cite{rvm}                        & 18.482                              & 0.417                                & 0.883                             & 0.876                      \\ [0.4 ex] 
Surface-SOS (w/ RVM)                  & 16.828 ($\downarrow$ 1.654)         & 0.356 ($\downarrow$ 0.061)           & 0.895 ($\uparrow$ 0.012)          & 0.891 ($\uparrow$ 0.015)    \\ [0.4 ex] 
\midrule
Scene \textit{Kevin}                  & SAD $\downarrow$                    & MSE $\downarrow$                     & mIoU $\uparrow$                   & Acc.$\uparrow$ \\ [0.35 ex] 
\midrule
Surface-SOS (w/o mask Init.)          & 13.336 (-)                          & 0.396 (-)                            & 0.836 (-)                         & 0.941 (-)  \\ [0.4 ex] 
Mask-RCNN \cite{maskrcnn}             & 16.012                              & 0.417                                & 0.824                             & 0.872 \\ [0.4 ex] 
Surface-SOS (w/ Mask-RCNN)            & 14.702 ($\downarrow$ 1.311)         & 0.411 ($\downarrow$ 0.006)           & 0.846 ($\uparrow$ 0.022)          & 0.879 ($\uparrow$ 0.007) \\ [0.4 ex] 
SAM \cite{segany}                     & 10.603                              & 0.386                                & 0.861                             & 0.924 \\ [0.4 ex]
Surface-SOS (w/ SAM)                  & 9.593 ($\downarrow$ 1.010)          & 0.356 ($\downarrow$ 0.030)           & 0.879 ($\uparrow$ 0.018)          & 0.932 ($\uparrow$ 0.008) \\ [0.4 ex] 
RVM \cite{rvm}                        & 11.366                              & 0.394                                & 0.860                             & 0.960   \\ [0.4 ex] 
Surface-SOS (w/ RVM)                  & \textbf{8.965 ($\downarrow$ 2.401)} & \textbf{0.362 ($\downarrow$ 0.032)}  & \textbf{0.881 ($\uparrow$ 0.021)} & \textbf{0.974 ($\uparrow$ 0.014) }  \\ [0.4 ex]  
\bottomrule[1pt]
\end{tabular}
\vspace{-5mm}
\label{tab:mask}
\end{table*}

\begin{figure}[htbp!]
\centering
\includegraphics[width=0.5\textwidth]{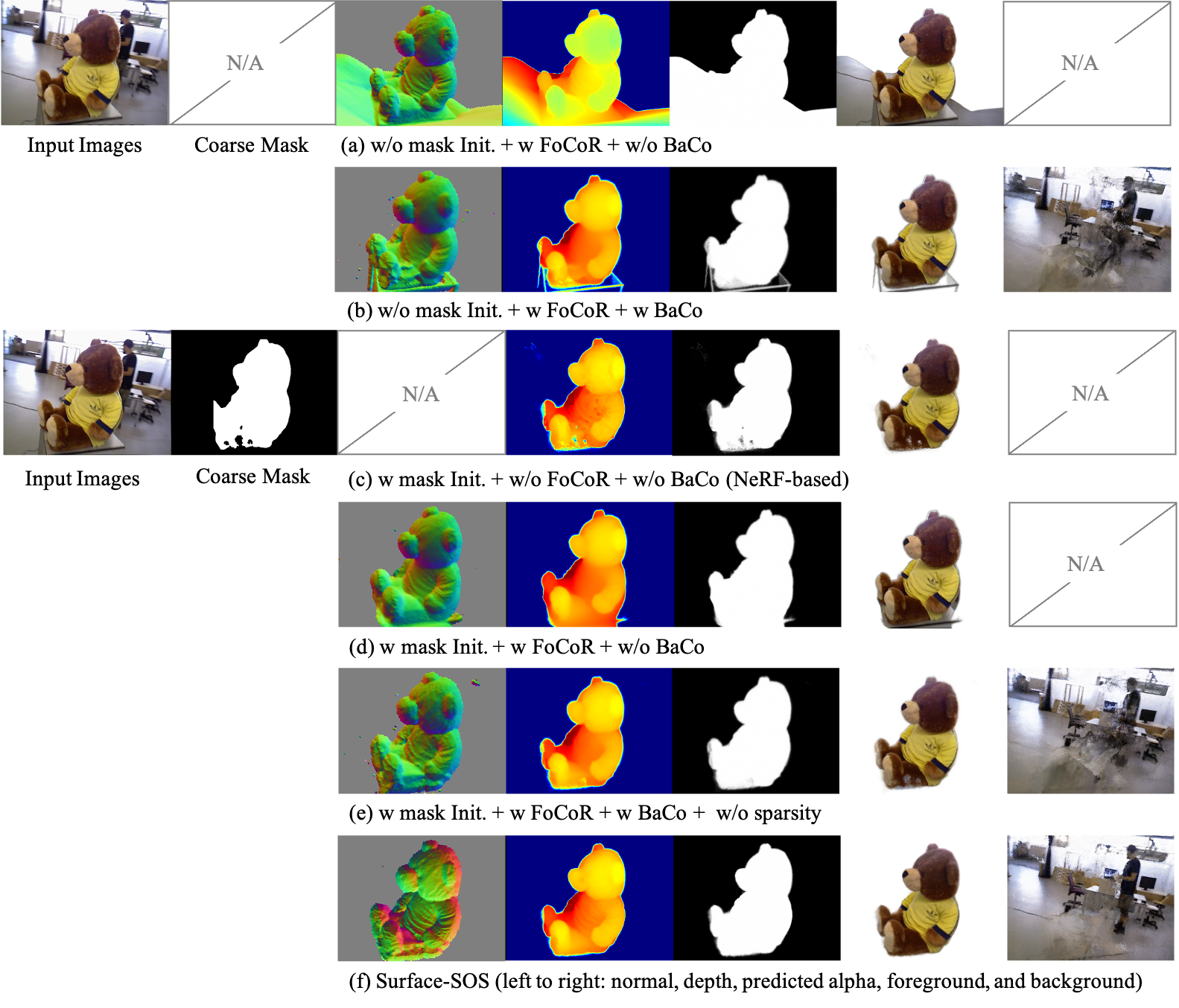} 
\vspace{-6mm}
\caption{Ablation Studies. Compare Surface-SOS to different design choices: with and without a coarse mutilated mask input, with a coarse mask and with or without our proposed FoCoR and BaCo modules, as well as the sparsity loss. Without the coarse mask initiation, Surface-SOS can decompose reasonable foreground and background, part of the desk is segmented out due to the view-consistent geometry for the static foreground. When providing the FoCoR module with coarse masks, the network is able to learn 3D geometry implicitly and generate an accurate foreground decomposition, By adding the background learning module (i.e., BaCo) and sparsity loss to the SDF-based surface representation, the resulting geometric surfaces become more compact and prevent holes in the object alpha matte.}
\label{ablation}
\end{figure}

To achieve a high-quality surface representation and analyze the correlation between the object surface representation and object segmentation, we present the performance with different design choices. These include variations with and without a coarse mutilated mask input, with a coarse mask, and with or without our proposed FoCoR and BaCo modules, as well as the inclusion of the sparsity loss. The corresponding performances are reported in Table~\ref{tab:ablation} and Fig.~\ref{ablation}. These results convey several observations. 

Firstly, Surface-SOS can effectively recover dense 3D surface structures from multi-view images even without auxiliary inputs of object masks. As we can see in Fig.~\ref{ablation}(a), this approach results in a reasonable foreground and background decomposition. 

Secondly, when providing the SDF-based surface representation (i.e., FoCoR module) with coarse masks, the network learns 3D geometry implicitly and generates an accurate foreground decomposition. Specifically, the background learning module (i.e., BaCo) prevents the occurrence of uncontrollable free surfaces, and the sparsity loss encourages the model to render images with the minimum content required for recovery and prevents the duplication of foreground representations, which further promotes the creation of an accurate foreground and background decomposition with fine detail. These examples demonstrate the benefits of accurately predicting object geometry using two complementary neural representations for self-supervised object segmentation.

\begin{figure}[htbp!]
\centering
\includegraphics[width=0.485\textwidth]{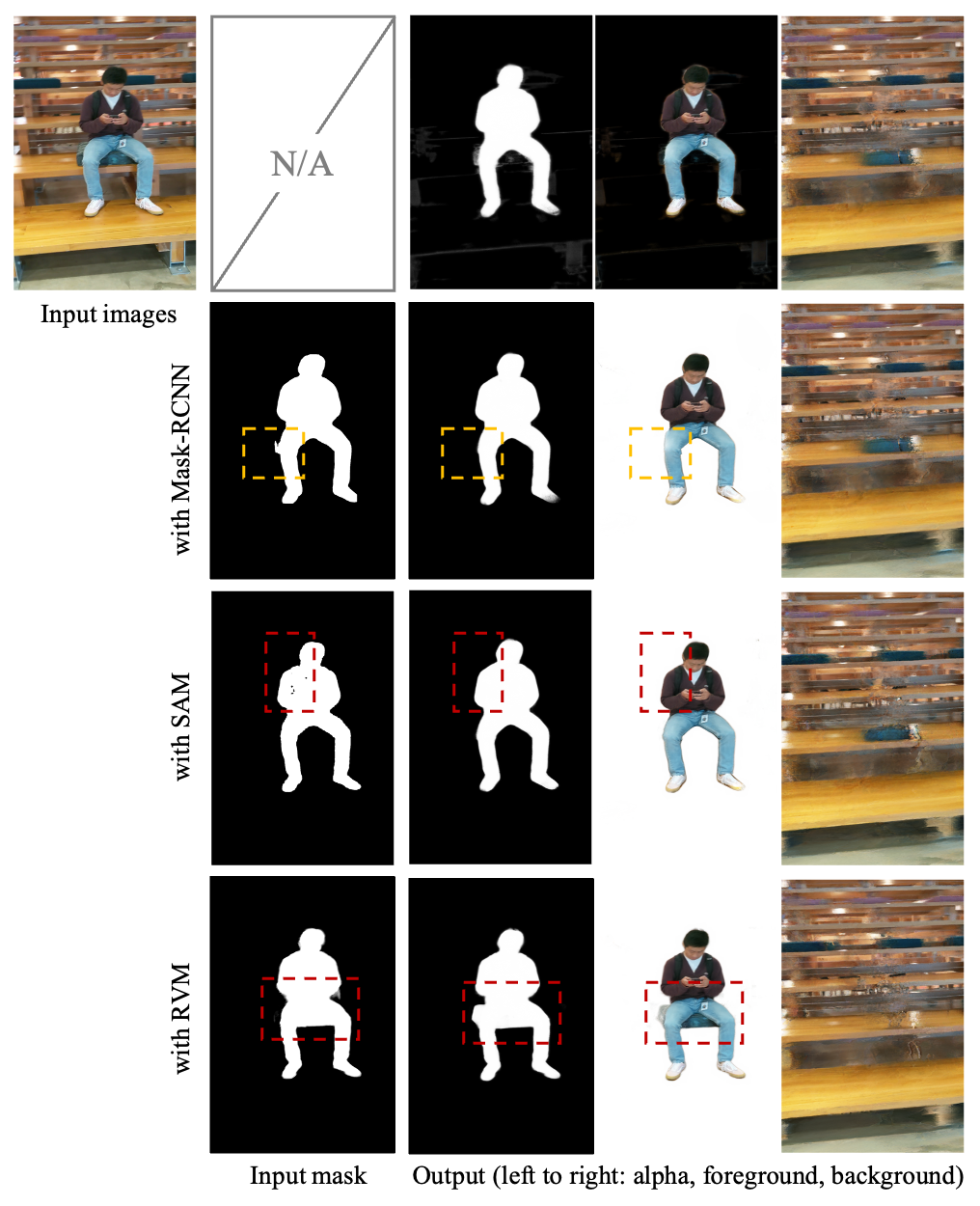} 
\vspace{-6mm}
\caption{Impact of the mask initialization. Surface-SOS can effectively decompose a complex scene into foreground and background without initiating a rough mask. By introducing coarse segmentation masks as additional input, Surface-SOS is able to refine single-view segmentation, such as Mask-RCNN \cite{maskrcnn}, SAM \cite{segany}, and RVM \cite{rvm}. For instance, compared to the coarse masks of Mask-RCNN and SAM, the initial mask of RVM contains incomplete cushions, Surface-SOS can effectively recover dense 3D surface structures from multi-view images and produce high-quality segmentation maps. These examples show that even rough segmentation results in this step can yield high-quality foreground at the end, making the system practically applicable. }
\label{abl_mask} 
\end{figure}

Moreover, by introducing coarse masks as additional input, Surface-SOS is able to refine the segmentation remarkably. In Fig.~\ref{abl_mask} we present an analysis of mask initialization in our framework by removing the coarse mask input, as well as applying several rough segmentation acquired by different single-view methods \cite{maskrcnn, segany, rvm}. Extensive experiments are presented in Table~\ref{tab:mask}, and the results clearly demonstrate that Surface-SOS outperforms all of the original single-view methods by a large margin. For instance, for the scenes \textit{Kevin}, in terms of SAD, MSE, mIoU, and Acc., Surface-SOS with the RVM masks initialization surpasses the RVM by -2.401, -0.032, 0.021 and 0.014, respectively. These examples show that even rough segmentation results in this step can yield high-quality foreground at the end, making the system practically applicable.

\section{Conclusion}

In this paper, we present Surface-SOS, a new self-supervised learning framework for delicate segmentation from multi-view images that are geometrically consistent. To leverage 3D object-level geometry and 2D image appearance cues of the one-to-one dense mapping in 3D space, we designed a special neural scene decomposition approach containing two complementary neural representation modules, i.e. FoCoR and BaCo, processing the foreground and background, respectively. In this manner, we can effectively decompose scenes into foreground and background, including its convincing segmentation maps. Our framework can be implemented to refine 2D single-view object segmentation on complex scenes with only unlabeled multi-view images. Extensive experiments on various multi-view benchmark datasets and monocular stereo videos validated the effectiveness of the Surface-SOS, significantly improving the supervised 2D single-view object segmentation results, and generating finer-grained segmentation than existing multi-view NeRF-based frameworks.

\begin{figure}[htbp!]
\centering
\includegraphics[width=0.495\textwidth]{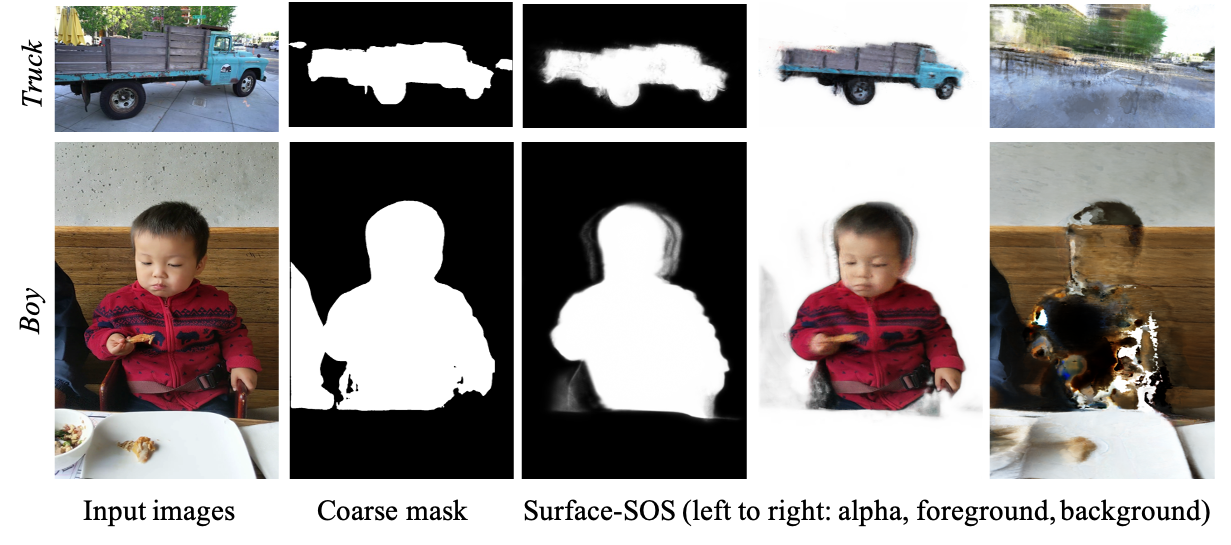} 
\vspace{-6mm}
\caption{Failure cases. On the unbounded scene \textit{Truck} from hand-held 360 capture Tank and Temples dataset \cite{tanks}, the generated results are blurry and lack fine details. On the custom stereo video \textit{Boy} from \cite{consistent}, it mistakenly matches several discrete pixel patches.}
\label{bad_case}
\end{figure}

Though promising, there are still some limitations and drawbacks of the proposed method. As we extract geometric constraints by leveraging the SDF-based surfaces representation from a sparse set of images, it cannot segment across scenes. Furthermore, it faces more challenges on unbounded scenes, such as the hand-held 360 capture large-scale Tank and Temples datasets \cite{tanks}, due to the lack of solid geometry in the scenes. Our method supports videos containing moderate object motion. It breaks for extreme object motion. Some failure cases are shown in Fig. \ref{bad_case}. Integrating our approach with learning-based pose estimation and neural dynamic representation of large motion and deformed objects is an interesting future direction.

\section*{Acknowledgments}
This work is financially supported for Outstanding Talents Training Fund in Shenzhen, Shenzhen Science and Technology Program-Shenzhen Cultivation of Excellent Scientific and Technological Innovation Talents project (Grant No. RCJC20200714114435057), Shenzhen Science and Technology Program-Shenzhen Hong Kong joint funding project (Grant No. SGDX20211123144400001), National Natural Science Foundation of China (Grant No. U21B2012), and MIGU-PKU META VISION TECHNOLOGY INNOVATION LAB. Jianbo Jiao is supported by the Royal Society grants IES\textbackslash R3\textbackslash223050 and SIF\textbackslash R1\textbackslash231009.


\bibliographystyle{IEEEtran}
\bibliography{citation}


 

\begin{IEEEbiography}[{\includegraphics[width=1in,height=1.25in,clip,keepaspectratio]{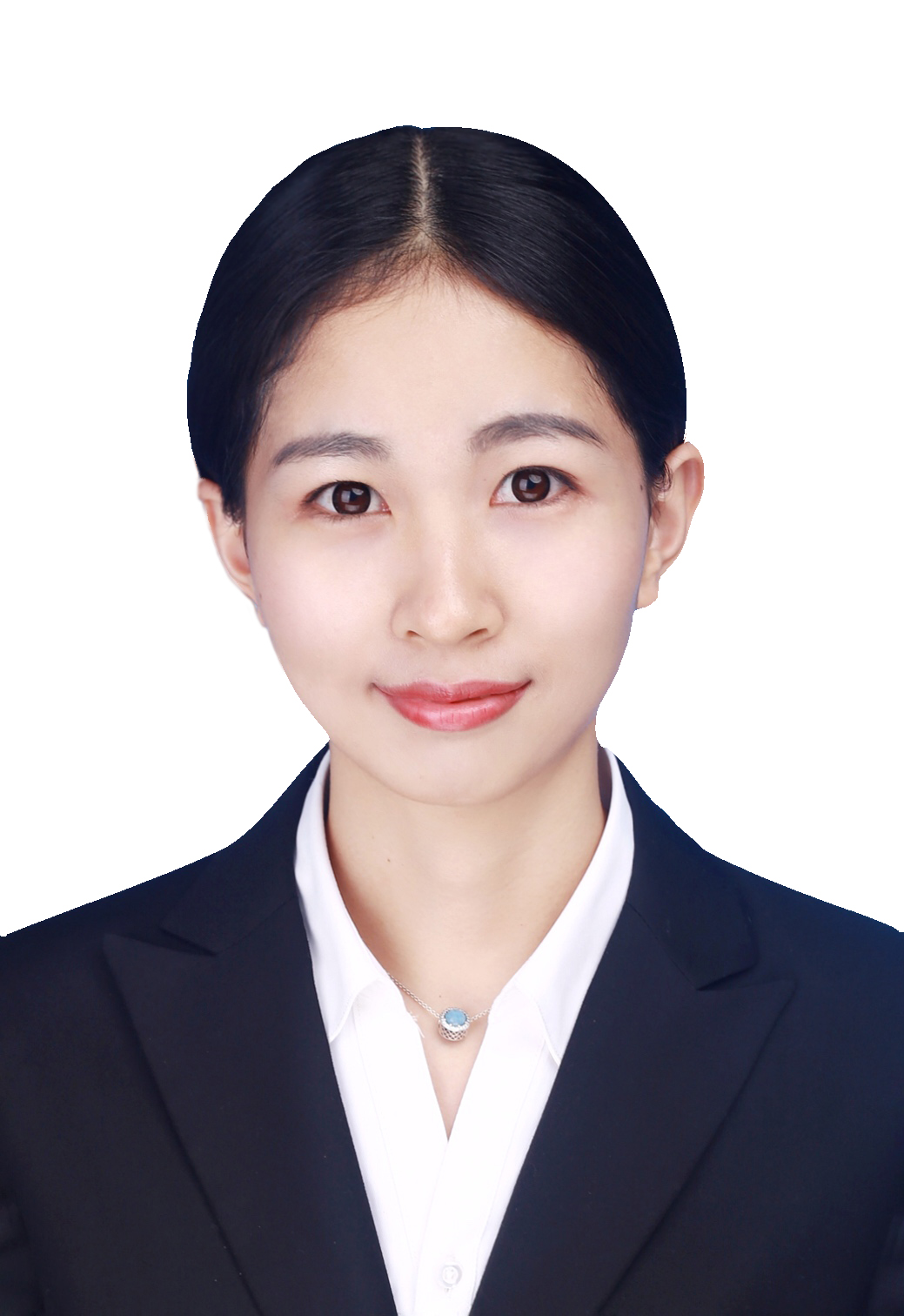}}]{Xiaoyun Zheng}
received the M.S. degree in Mechanical Engineering from Tongji University, Shanghai, China, in 2019. She is currently pursuing the Ph.D. degree with the School of Computer Science, Peking University Shenzhen Graduate School, Shenzhen, China. Her research interests include video/image processing, computer vision, and 3D human reconstruction, etc.
\end{IEEEbiography}

\begin{IEEEbiography}[{\includegraphics[width=1in,height=1.25in,clip,keepaspectratio]{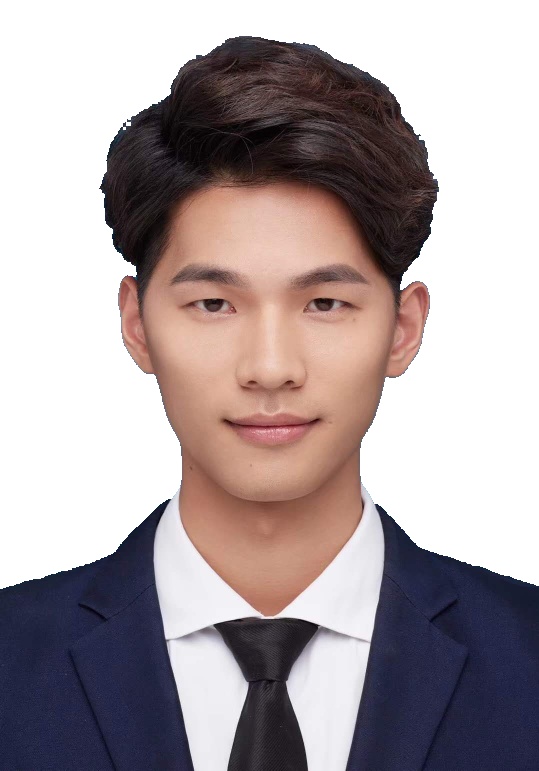}}]{Liwei Liao}
received the M.S. degree in Micro Electronics from Peking University, Beijing China, in 2019. He is currently pursuing the Ph.D. degree in Computer Science, Peking University Shenzhen Graduate School, Shenzhen, China. His research interests include multi-modal machine learning and 3D human reconstruction, etc.
\end{IEEEbiography}

\begin{IEEEbiography}[{\includegraphics[width=1in,height=1.25in,clip,keepaspectratio]{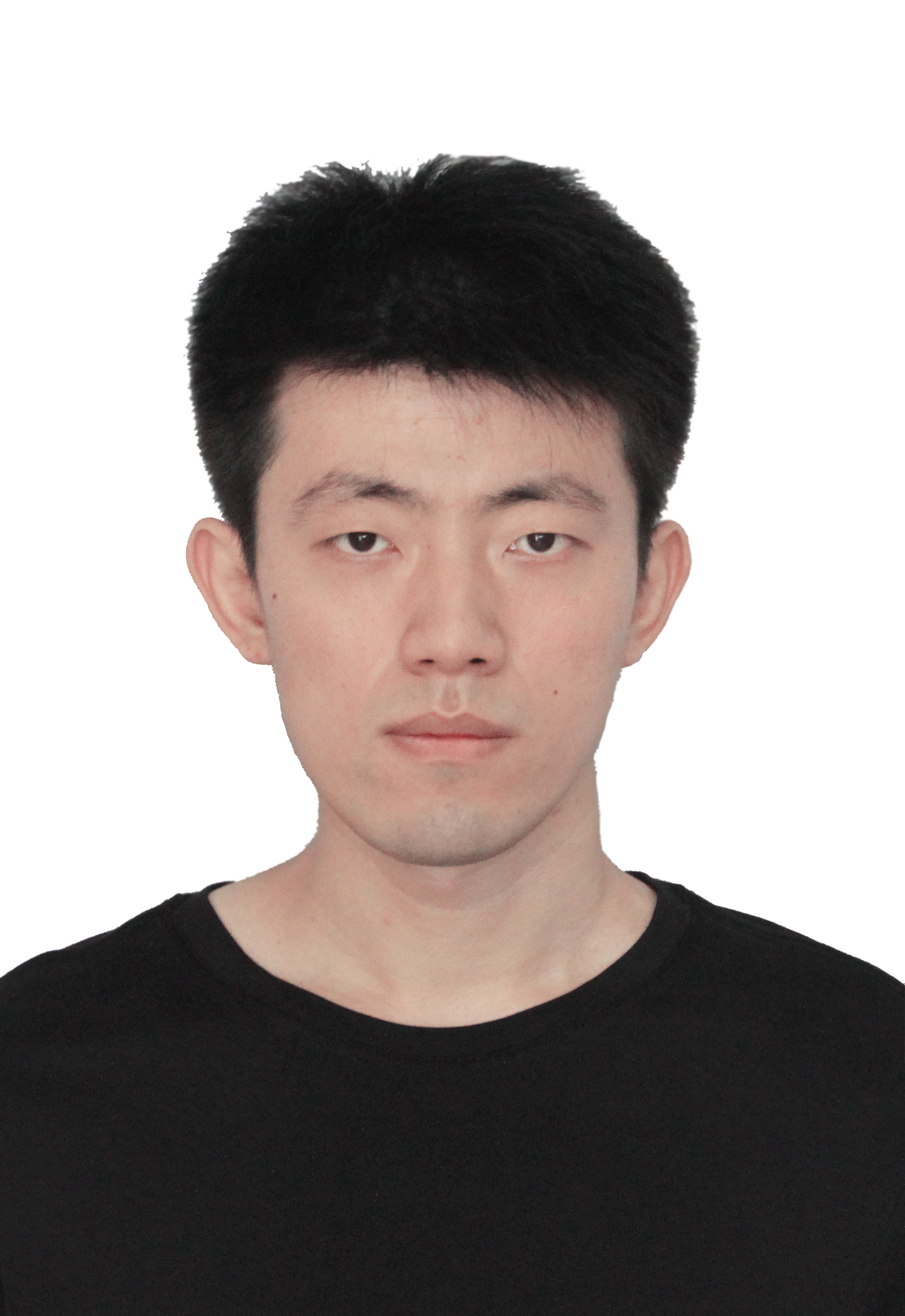}}]{Jianbo Jiao}
(Member, IEEE) received the PhD degree in computer science from the City University of Hong Kong, in 2018. He is currently an Assistant Professor in the School of Computer Science at the University of Birmingham, a Royal Society Short Industry Fellow, and a Visiting Researcher at the University of Oxford, United Kingdom. Before joining Birmingham, he was a Postdoctoral Researcher in the Department of Engineering Science at the University of Oxford. He was the recipient of the Hong Kong PhD Fellowship Scheme (HKPFS). He was a Visiting Scholar with the Beckman Institute at the University of Illinois at Urbana-Champaign from 2017 to 2018. His research interests include machine learning and computer vision.
\end{IEEEbiography}

\begin{IEEEbiography}[{\includegraphics[width=1in,height=1.25in,clip,keepaspectratio]{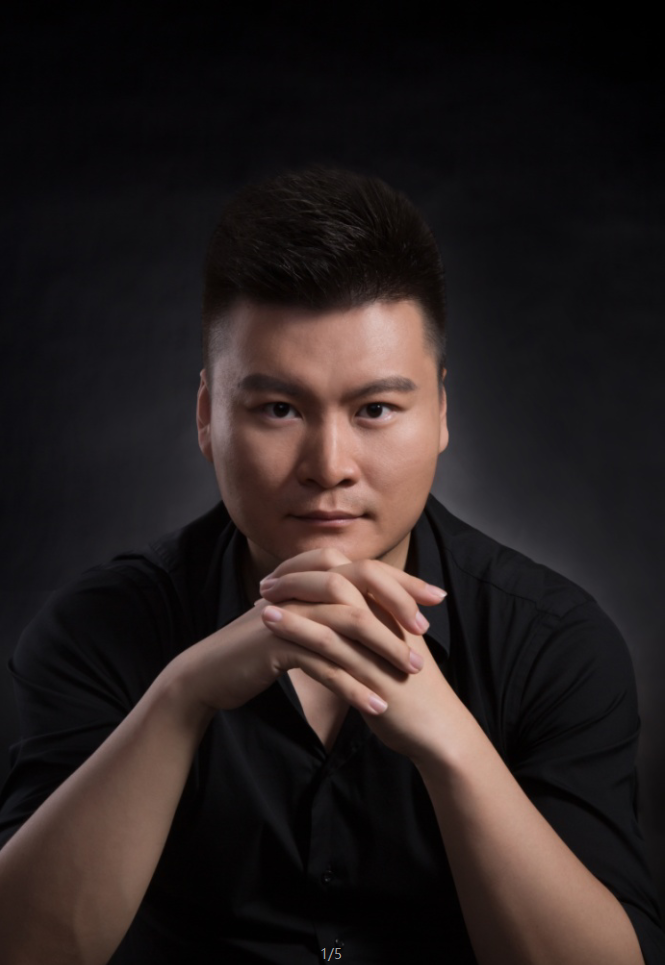}}]{Feng Gao}
received his B.S. degree in Computer Science from University College London in 2007, and Ph.D. degree in Computer Science from Peking University in 2018. He was a post-doctoral research fellow at the Future Laboratory,Tsinghua University, from 2018 to 2020. He joins Peking University as Assistant Professor since 2020. His research interesting is working on the intersection of Computer Science and Art, including but not limit on artificial intelligence and painting art, deep learning and painting robot, etc.
\end{IEEEbiography}

\begin{IEEEbiography}[{\includegraphics[width=1in,height=1.25in,clip,keepaspectratio]{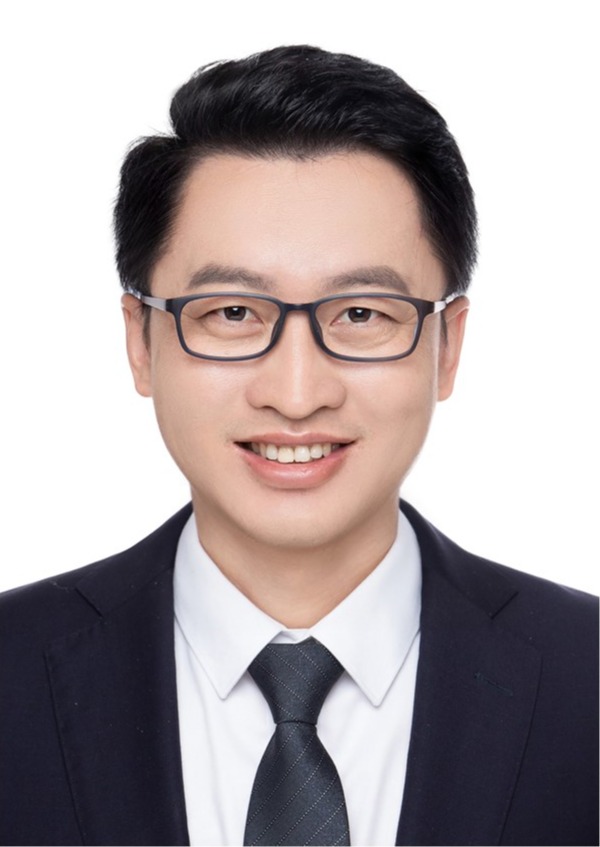}}]{Ronggang Wang}
(Member, IEEE) received the Ph.D. degree from the Institute of Computing Technology, Chinese Academy of Sciences, in 2006. He is currently a Professor with the School of Electronic and Computer Engineering, Peking University Shenzhen Graduate School. His research interests include immersive video coding and processing. He has made over 100 technical contributions to ISO/IEC MPEG, IEEE 1857 and China AVS. He has authored more than 150 papers and held more than 100 patents. He has been serving as the IEEE 1857.9 Immersive video coding standard sub-group Chair and China AVS virtual reality sub-group Chair since 2016.
\end{IEEEbiography}

\vspace{11pt}


\vfill

\end{document}